\documentclass[10pt,journal,compsoc]{IEEEtran}

\ifCLASSOPTIONcompsoc
\usepackage[nocompress]{cite}
\else
\usepackage{cite}
\fi

\usepackage{times}
\usepackage{graphicx}
\usepackage{amsmath}
\usepackage{amsthm}
\usepackage{amssymb}
\usepackage{mathtools}
\usepackage{color}
\usepackage{array}
\usepackage[pagebackref=true,breaklinks=true,colorlinks,bookmarks=false]{hyperref}
\usepackage{setspace}
\usepackage{subfig}
\usepackage[english]{babel}
\newcommand{\minesubsec}[1]{\vspace{0.3cm} \noindent{\bf #1~}}

\newcommand{\comment}[1]{}

\newcommand{\BQ}{\begin{equation}}
\newcommand{\EQ}{\end{equation}}

\DeclareMathOperator*{\argmin}{argmin}

\def\Ninf{{N \rightarrow +\infty}}
\def\limitN{{\lim\limits_\Ninf}}
\def\intp{{\int\limits_{p=-M}^{M}}}
\def\intq{{\int\limits_{q=-M}^{M}}}

\def\mb{\overline{m}}

\newtheorem{lemma}{Lemma}

\newtheorem{theorem}{Theorem}

\newcommand{\ignore}[1]{}

\begin{document}
	
	\title{Best-Buddies Similarity - Robust Template Matching using Mutual Nearest Neighbors}

	\author{Shaul~Oron, Tali~Dekel, Tianfan~Xue, William~T.~Freeman, Shai~Avidan		
		\IEEEcompsocitemizethanks{\IEEEcompsocthanksitem S. Oron, Department of Electrical Engineering, Tel-Aviv University
		\protect\\
		E-mail: shauloro@post.tau.ac.il
		\IEEEcompsocthanksitem T. Dekel, MIT Computer Science and Artificial Intelligence Lab, Google
		\protect\\
			E-mail: tdekel@google.com
			\IEEEcompsocthanksitem T. Xue, MIT Computer Science and Artificial Intelligence Lab
		\protect\\
			E-mail: tfxue@mit.edu			
			\IEEEcompsocthanksitem W.T. Freeman, MIT Computer Science and Artificial Intelligence Lab, Google
		\protect\\
			E-mail: billf@mit.edu
			\IEEEcompsocthanksitem S. Avidan, Department of Electrical Engineering, Tel-Aviv University
		\protect\\
		E-mail: avidan@eng.tau.ac.il
			}
		}
	
	\maketitle
	
\begin{abstract}	
We propose a novel method for template matching in unconstrained environments. Its essence is the Best-Buddies Similarity (BBS), a useful, robust,  and parameter-free similarity measure between two sets of points. 
BBS is based on counting the number of Best-Buddies Pairs (BBPs)---pairs of points in source and target sets, where each point is the nearest neighbor of the other.
 BBS has several key features that make it robust against complex geometric deformations and high levels of outliers, such as those arising from background clutter and occlusions. We study these properties, provide a statistical analysis that justifies them, and demonstrate the consistent success of BBS on a challenging real-world dataset while using different types of features.
\end{abstract}

\section{Introduction}
Finding a template patch in a target image is a core component in a variety of computer vision applications such as object detection, tracking, image stitching and 3D reconstruction. In many real-world scenarios, the template---a bounding box containing a region of interest in the source image ---undergoes complex deformations in the target image: the  background can change and the object may undergo nonrigid deformations and partial occlusions.

\begin{figure}[t]
	\centering
	\centering
	\includegraphics[width=1.0\linewidth, angle =0]{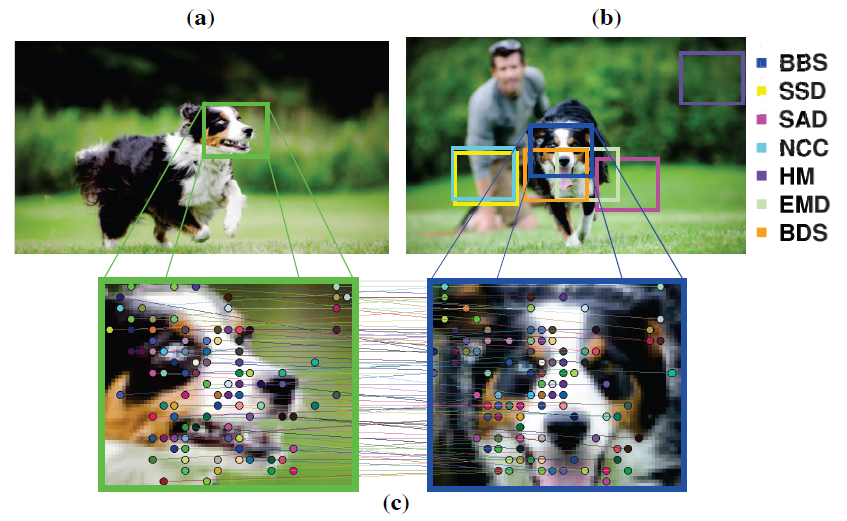}
	
		\caption{\hspace{-0.03cm}{\bf Best-Buddies Similarity (BBS) for Template Matching:} (a), The template, marked in green, contains an object of interest against a background. (b), The object in the target image undergoes complex deformation (background clutter and large geometric deformation); the detection results using different similarity measures are marked on the image (see legend); our result is marked in blue. (c), The Best-Buddies Pairs (BBPs) between the template and the detected region are mostly found the object of interest and not on the background; each BBP is connected by a line and marked in a unique color. }\label{fig:teaser}
		\vspace{-0.2cm}
\end{figure}

Template matching methods have been used with great success over the years but they still suffer from a number of drawbacks.
Typically, all pixels (or features) within the template and a candidate window in the target image are taken into account when measuring their similarity.  This is undesirable in some cases, for example, when the background behind the object of interest changes between the template and the target image (see Fig.~\ref{fig:teaser}).  In such cases, the dissimilarities between pixels from different backgrounds may be arbitrary, and accounting for them may lead to false detections of the template  (see Fig.~\ref{fig:teaser}(b)).

 In addition, many template matching methods assume a specific parametric deformation model between the template and the target image (e.g., rigid, affine transformation, etc.). This limits the type of scenes that can be handled, and may require estimating a large number of parameters when complex deformations are considered.

In order to address these challenges, we introduce a novel similarity measure termed \emph{Best-Buddies Similarity (BBS)}, and show that it can be applied successfully to template matching \emph{in the wild}. In order to compute the BBS we first represent both the template patch and candidate query patches as point sets in $\mathbb{R}^d$. Then, instead of searching for a parametric deformation between template and candidate we directly measure the similarity between these point sets. We analyze key features of BBS, and perform extensive evaluation of its performance compared to a number of commonly used alternatives on challenging datasets.  

BBS measures the similarity between two sets of points in $\mathbb{R}^d$. A key feature of this measure is that it relies only on a subset (usually small) of pairs of points -- the \emph{Best-Buddies Pairs (BBPs)}. A pair of points is considered a BBP if the points are mutual nearest neighbors, i.e. each point is the nearest neighbor of the other in the corresponding point set. BBS is then taken to be the fraction of BBPs out of all the points in the set.

Albeit simple, this measure turns out to have important and nontrivial properties. Because BBS counts only the pairs of points that are best buddies, it is robust to significant amounts of outliers. Another, less obvious property is that the BBS between two point sets is maximal when the points are drawn from the same distribution, and drops sharply as the distance between the distributions increases. In other words, if two points are BBP, they were likely drawn from the same distribution. We provide a statistical formulation of this observation, and analyze it numerically in the 1D case for point sets drawn from distinct Gaussian distributions (often used as a simplified model for natural images). 

Modeling image data as distributions, i.e. using histograms, was successfully applied to many computer vision tasks, due to its simple yet effective non-parametric representation. A prominent distance measure between histograms is the Chi-Square ($\chi^2$) distance, in which contributions of different bins, to the similarity score, are proportional to the overall probability stored in those bins.

In this work we show that for sufficiently large sets, BBS converges to the $\chi^2$ distance between distributions. However, unlike $\chi^2$ computing BBS is done directly on the raw data without the need to construct histograms. This is advantageous as it alleviates the need to choose the histogram bin size. Another benefit is the ability to work  with high dimensional representation, such as Deep features, for which constructing histograms is not tractable. 

More generally, we show a link between BBS and a well known statistical measure. This provides additional insight into the statistical properties of mutual nearest neighbors, and also sheds light on the ability of BBS to reliably match features coming from the same distribution, in the presence of outliers.

We apply the BBS measure to template matching by representing both the template and each of the candidate image regions as point sets in a joint location-appearance space. To this end, we use normalized coordinates for location and experiment with both color as well as Deep features for appearance (although, BBS is not restricted to these specific choices). BBS is used to measure the similarity between the two sets of points in these spaces. The aforementioned properties of BBS now readily apply to template matching. That is, pixels on the object of interest in both the template and the candidate patch can be thought of as originating from the same underlying distribution. These pixels in the template are likely to find best buddies in the candidate patch, and hence would be considered as inliers. In contrast, pixels that come from different distributions, e.g., pixels from different backgrounds, are less likely to find best buddies, and hence would be considered outliers (see Fig.~\ref{fig:teaser}(c)). Given this important property, BBS bypasses the need to explicitly model the underlying object appearance and deformation.

	To summarize, the main contributions of this paper are: (a) introducing BBS --   a useful, robust, parameter-free measure for template matching in unconstrained environments, (b) analysis providing theoretical justification of its key features and linking BBS with the Chi-Square distance, and (c) extensive evaluation on challenging real data, using different feature representations, and comparing BBS to a number of commonly used template matching methods. A preliminary version of this paper appeared in CVPR 2015~\cite{Dekel2015BBS}.
	\textit{}\begin{figure}[t!]
	\centering
	\begin{tabular}{c}
		\includegraphics[width = 0.95\columnwidth]{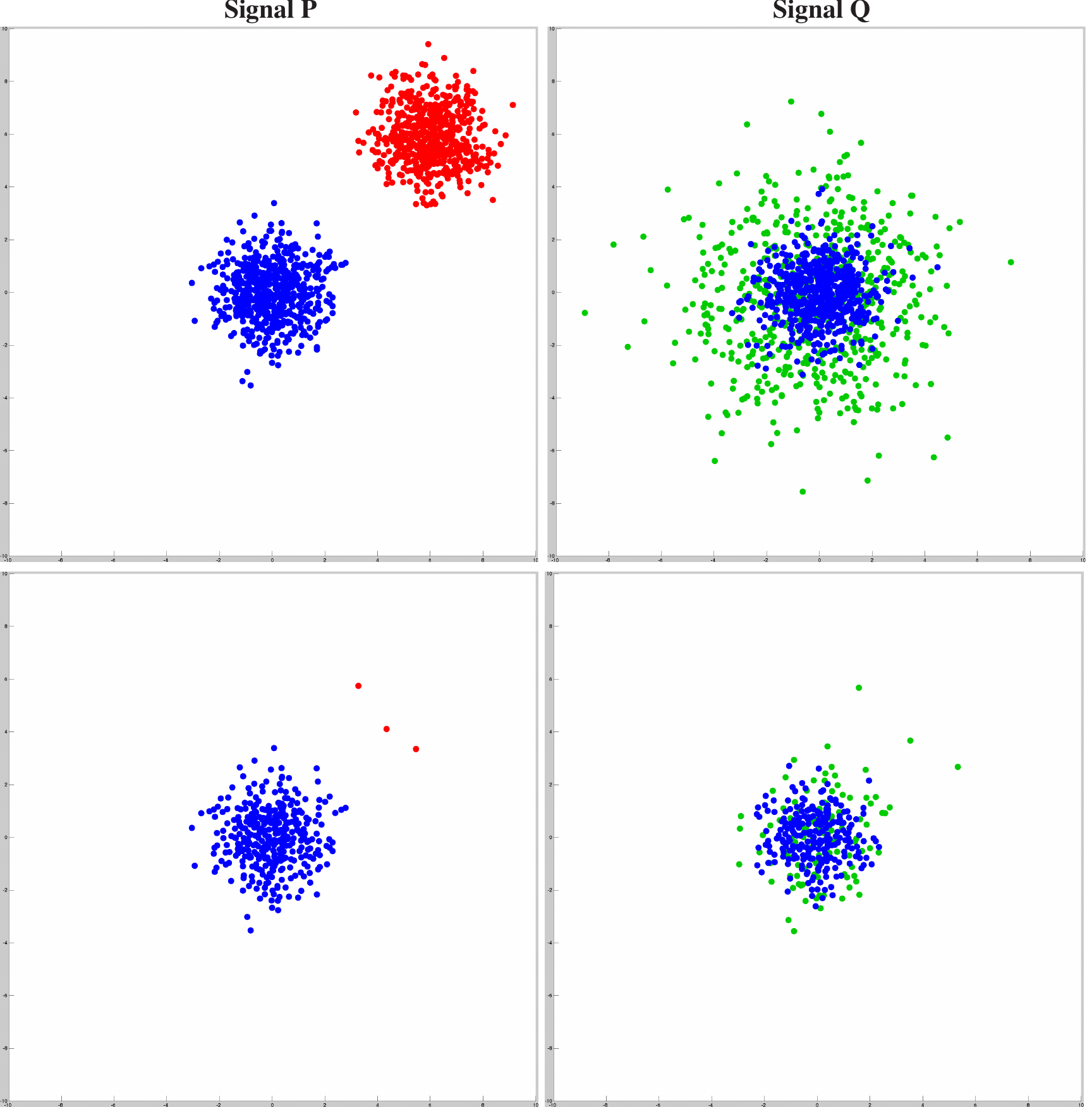} \\
	\end{tabular}\hspace{-0.02cm}
	\caption{\hspace{-0.02cm}{\bf Best-Buddies
Pairs (BBPs) between 2D Gaussian Signals:} First row, Signal $P$ consists of ``foreground'' points drawn from a normal distribution,  $N(\mu_1,\sigma_1)$, marked in blue; and ``background'' points drawn from $N(\mu_2,\sigma_2)$, marked in red. Similarly, the points in the second signal Q are drawn from the same distribution $N(\mu_1,\sigma_1)$, and a different background distribution $N(\mu_3,\sigma_3)$. The color of points is for illustration only, i.e., BBS does not know which point belongs to which distribution. Second row, only the BBPs between the two signals which are mostly found between foreground points. }
		\label{fig:bb_points_2Dgauss} \vspace{-0.5cm}
\end{figure}
\section{Related Work}
	Template matching algorithms depend heavily on the similarity measure used to match the template and a candidate window in the target image.
Various similarity measures have been used for this purpose. The most popular are the Sum of Squared Differences (SSD), Sum of Absolute Differences (SAD) and  Normalized Cross-Correlation (NCC), mostly due to their computational efficiency \cite{ouyang2012performance}.
 Different variants of these measures have been proposed to deal with illumination changes and noise \cite{Hel-OrHD14,elboher2013asymmetric}. 

Another family of measures is composed of robust error functions such as M-estimators \cite{chen2003fast, sibiryakov2011fast} or Hamming-based distance \cite{shin2007fast, pele2008robust}, which are less affected by additive noise and 'salt and paper' outliers than cross correlation related methods.  However, all the methods mentioned so far assume a strict rigid geometric deformation (only translation) between the template and the target image, as they penalize pixel-wise differences at corresponding positions in the template and the query region. 

A number of methods extended template matching to deal with parametric transformations (e.g., \cite{tsai2002rotation,kim2007grayscale}). Recently, Korman {\em et al.} \cite{cvpr2013Fast_Match} introduced a template matching algorithm under 2D affine transformation that  guarantees  an approximation to the globally optimal solution. Likewise, Tian and Narasimhan \cite{tian2012globally} find a globally optimal estimation of nonrigid image distortions. However,  these methods assume a one-to-one mapping between the template and the query region for the underlying transformation.  Thus, they are prone to errors in the presence of many outliers, such as those caused by occlusions and background clutter. Furthermore, these methods assume a parametric model for the distortion geometry, which is not required in the case of BBS. 

Measuring the similarity between color histograms, known as Histogram Matching (HM), offers a non-parametric technique for dealing with deformations and is commonly used in visual tracking \cite{comaniciu2000real,perez2002color}. 
Yet, HM completely disregards geometry, which is a powerful cue. Further, all pixels are evenly treated. Other tracking methods have been proposed to deal with cluttered environments and partial occlusions \cite{Bao12,Jia12}. But unlike tracking, we are interested in detection in a single image, which lacks the redundant temporal information given in videos. 

Olson \cite{Olson02} formulated template matching in terms of maximum likelihood estimation, where an image is represented in a 3D location-intensity space.
Taking this approach one step further, Oron {\em et al.}\cite{Oron2014LOT} use $xyRGB$ space and reduced template matching to measuring the EMD \cite{Rubner00} between two point sets. Unlike EMD, BBS does not require $1:1$ matching. It therefore does not have to account for all the data when matching, making it more robust to outliers. 

The BBS is a bi-directional measure. The importance of such two-side agreement has been demonstrated by the Bidirectional similarity (BDS) in \cite{Simakov2008summarizing} for visual summarization. Specifically, the BDS was used as a similarity measure between two images, where an image is represented by a set of patches. The BDS sums over the  distances between each patch in one image to its nearest neighbor in the other image, and vice versa. 

In the context of image matching, another widely used measure is the Hausdorff distance \cite{huttenlocher1993comparing}. 
To deal with occlusions or degradations, Huttenlocher {\em et al.} \cite{huttenlocher1993comparing} proposed a fractional Hausdorff distance in which the $\text{K}^{th}$ farthest point is taken instead of the most farthest one.  Yet, this measure highly depends on $K$ that needs to be tuned. Alternatively, Dubuisson and Jain \cite{Dubuisson94} replace the max operator with sum, which is similar to the way BDS is defined.

In contrast, the BBS is based on a \emph{count} of the BBPs, and makes only implicit use of their actual distance. Moreover, the BDS  does not distinguish between inliers and outliers. These proprieties makes the BBS a more robust and reliable measure as demonstrated by our experiments.

We show a connection between BBS and the Chi-Square ($\chi^2$) distance used as a distance measure between distributions (or histograms). Chi-Square distance comes from the $\chi^2$ test-statistic ~\cite{snedegor1967statistical} where it is used to test the fit between a distribution and observed frequencies. $\chi^2$ was successfully applied to a wide range of computer vision tasks such as texture and shape classification ~\cite{Varma09,belongie2002shape}, local descriptors matching ~\cite{forssen2007shape}, and boundary detection ~\cite{martin2004learning} to name a few. 

It is worth mentioning, that the term {\em Best Buddies} was used by Pomeranz {\em et al.} \cite{PomeranzSB11} in the context of solving jigsaw puzzles. Specifically, they used a  metric similar to ours in order to determine if a pair of pieces are compatible with each other.

The power of mutual nearest neighbors was previously leveraged for tasks such as image matching~\cite{Li2015}, classification of images~\cite{Liu2010} and natural language data~\cite{Ozaki2011}, clustering ~\cite{Hu2012} and more. In this work we demonstrate its use for template matching while providing some new statistical analysis.

	\begin{figure}[t!]
	\centering
	\includegraphics[width = 0.48\textwidth]{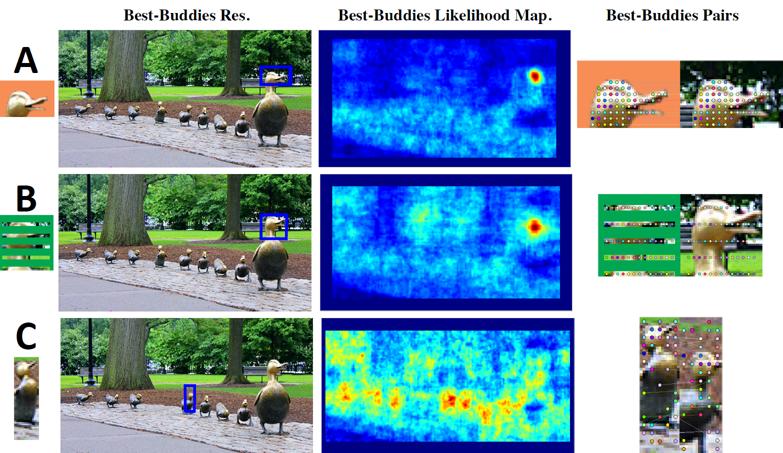} 
	\caption{{\bf BBS template matching results.} Three toys examples are shown: (A) cluttered background, (B) occlusions, (C) nonrigid deformation. The template (first column) is detected in the target image (second column) using the BBS; the  results using BBS are marked in a blue. The likelihood maps (third column) show well-localized distinct modes. The BBPs are shown in last column. See text for more details.}
		\label{fig:duck} \vspace{-0.1cm}
\end{figure}

\section{Best-Buddies Similarity} \label{sec:method}
\begin{figure}[t!]
  	\centering
  	\includegraphics[width = 0.48\textwidth]{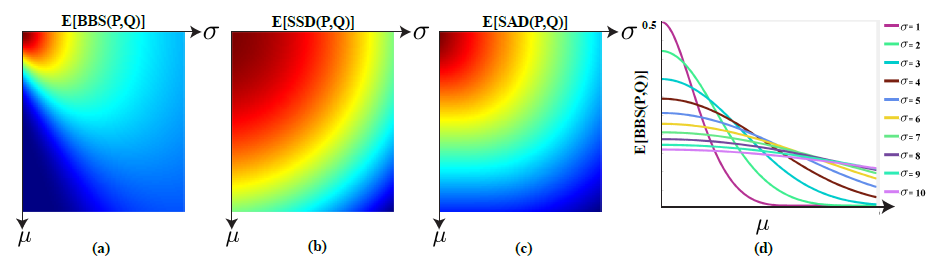} 
  	\caption{{\bf The expectation of BBS in the 1D Gaussian case:} Two point sets, P and Q, are generated by sampling points from $N(0,1)$, and $N(\mu,\sigma)$, respectively. (a), the approximated expectation of BBS(P,Q) as a function of $\sigma$ (x-axis), and $\mu$ (y-axis).(b)-(c), the expectation of SSD(P,Q), and SAD(P,Q), respectively. (d), the expectation of BBS as a function of $\mu$ plotted for different $\sigma$.}
  	\label{fig:theory} \vspace{-0.3cm}
  \end{figure}

Our goal is to match a template to a given image, in the presence of high levels of outliers (i.e., background clutter, occlusions) and nonrigid deformation of the object of interest. We follow the traditional sliding window approach and compute the Best-Buddies Similarity (BBS) between the template and every window (of the size of the template) in the image. In the following, we give a general definition of BBS and demonstrate its key features via simple intuitive toy examples. We then statistically analyze these features in Sec.~\ref{fig:theory}.  \vspace{-0.1cm}

\minesubsec{General Definition:}
BBS measures the similarity between two sets of points \mbox{$P\!=\!\{p_i\}_{i=1}^{N_P}$} and \mbox{$Q\!=\!\{q_i\}_{i=1}^{N_Q} $}, where  \mbox{$p_i, q_i \in \mathbb{R}^d$.}  The BBS is the fraction of \emph{Best-Buddies Pairs} (BBPs)  between the two sets. Specifically, a pair of points $\{ p_i \in P, q_j\in Q \}$ is a BBP if $p_i$ is the nearest neighbor of $q_j$  in the set $Q$, and vice versa. Formally, 
{\small \BQ  \label{eq: bb1}
bb(p_i,q_j,P,Q) = \left\{  
\begin{array}{ll}
	1 & {\small \text{NN}(p_i,Q)=q_j \land \text{NN}(q_j,P)=p_i}\\
	0 & \text{otherwise}
\end{array}
\right.\vspace{-0.02cm}\EQ} where, ${\small \text{NN}(p_i,Q)\!=\!\argmin\limits_{q\in Q}d(p_i,q)}$, and $d(p_i,q)$ is some distance measure. The BBS between the point sets $P$ and $Q$ is given by: \vspace{-0.04cm} 
\BQ \label{eq: BB} 
	\text{BBS}(P,Q) = \frac{ 1 }{ \min\{\text{$N_P,N_Q$}\} }\cdot \sum_{i=1}^{N_P}\sum_{j=1}^{N_Q} bb(p_i,q_j,P,Q).
\EQ
The key properties of the BBS are: 1) it relies only on a (usually small) subset of matches i.e.,  pairs of points that are BBPs, whereas the rest are considered as outliers. 2) BBS finds the bi-directional inliers in the data without any prior knowledge on the data or its underlying deformation. 3) BBS uses \emph{rank}, i.e., it counts the number of BBPs, rather than using the actual distance values. 

To understand why these properties are useful, let us consider a simple 2D case of two point sets $P$ and $Q$. The set $P$ consist of 2D points drawn from two different normal distributions, $N(\mu_1,\Sigma_1)$, and $N(\mu_2,\Sigma_2)$. Similarly, the points in $Q$ are drawn from the same distribution $N(\mu_1,\Sigma_1)$, and a different distribution  $N(\mu_3,\Sigma_3)$ (see first row in Fig.~\ref{fig:bb_points_2Dgauss}). The distribution $N(\mu_1,\Sigma_1)$ can be treated as a \emph{foreground} model, whereas $N(\mu_2,\Sigma_2)$ and $N(\mu_3,\Sigma_3)$ are two different \emph{background} models. As can be seen in Fig.~\ref{fig:bb_points_2Dgauss}, the BBPs are mostly found  between the foreground points in $P$ and $Q$. For set $P$, where the foreground and background points are well separated, $95\%$ of the BBPs are foreground points. For set $Q$, despite the significant overlap between foreground and background, $60\%$ of the BBPs are foreground points. 

This example demonstrates the robustness of BBS to high levels of outliers in the data. BBS captures the foreground points and does not force the background points to match. In doing so, BBS sidesteps the need to model the background/foreground parametrically or have a prior knowledge of their underlying distributions. This shows that a pair of points $\{p,q\}$ is more likely to be BBP if $p$ and $q$ are drawn from the same distribution. We formally prove this general argument for the 1D case in Sec.~\ref{sec:1d gauss}.  With this observations in hand, we continue with the use of BBS for template matching.

\subsection{BBS for Template Matching}
To apply BBS to template matching, one needs to convert each image patch to a point set in $\mathbb{R}^d$. Following \cite{Oron2014LOT}, we use a joint spatial-appearance space which was shown to be useful for template matching.  
BBS, as formulated in equation \eqref{eq: BB}, can be computed for any arbitrary feature space and for any distance measure between point pairs. In this paper we focus on two specific appearance representations: (i) using color features, and (ii) using Deep features taken from a pretrained neural net. Using such Deep features is motivated by recent success in applying features taken from deep neural nets to different applications ~\cite{Ma-ICCV-2015,Wang_2015_ICCV}. A detailed description of each of these feature spaces is given in Section \ref{sec:details}.

Following the intuition presented in the 2D Gaussian example (see Fig.~\ref{fig:bb_points_2Dgauss}), the use of BBS for template matching allows us to overcome several significant challenges such as background clutter,  occlusions, and nonrigid deformation of the object. This is demonstrated in three synthetic examples shown in Fig.~\ref{fig:duck}. The templates $A$ and $B$ include the object of interest in a cluttered background, and under occlusions, respectively. In both cases the templates are successfully matched to the image despite the high level of outliers. As can be seen, the BBPs are found only on the object of interest, and the BBS likelihood maps have a distinct mode around the true location of the template. In the third example, the template $C$ is taken to be a bounding box around the forth duck in the original image, which is removed from the searched image using inpating techniques. In this case, BBS matches the template to the fifth duck, which can be seen as a nonrigid deformed version of the template. Note that the BBS does not aim to solve the pixel correspondence. In fact, the BBPs are not necessarily  semantically correct (see third row in Fig.~\ref{fig:duck}), but rather pairs of points that likely originated from the same distribution.  This property, which we next formally analyze,  helps us deal with complex visual and geometric deformations in the presence of outliers.

	\section{Analysis} \label{sec:1d gauss}
So far, we have empirically demonstrated that the BBS is robust to outliers, and results in well-localized modes. In what follows, we give a statistical analysis that justifies these properties, and explains why using the count of the BBP is a good similarity measure. Additionally, we show that for sufficiently large sets BBS converges to the well known Chi-Square. This connection with $\chi^2$ provides additional insight into the way BBS handles outliers.

\subsection{Expected value of BBS}
We begin with a simple mathematical model in 1D, in which an ``image" patch is modeled as a set of points drawn from a general distribution. Using this model, we derive the expectation of BBS between two sets of points, drawn from two given distributions  $f_P(p)$ and $f_Q(q)$, respectively. We then analyze numerically the case in which $f_P(p)$, and $f_Q(q)$ are two different normal distributions. Finally, we relate these results to the multi-dimentional case.
We show that the BBS distinctively captures points that are drawn from similar distributions. That is, we prove that the likelihood of a pair of points  being BBP, and hence the expectation of the BBS, is maximal when the points in both sets are drawn from the same distribution, and drops sharply as the distance between the two normal distributions increases.

\minesubsec{One-dimentional Case:}
Following Eq.~\ref{eq: BB}, the expectation BBS(P,Q), over all possible samples of P and Q is given by:
\BQ
\label{eq:bb_E}
\begin{array}{ll}
	E[\text{BBS}(P,Q)]=&\frac{1}{\min\{N_P,N_Q\} }\sum\limits_{i=1}^{N_P} \sum\limits_{j=1}^{N_Q}E[ bb_{i,j}(P, Q)], \\
\end{array}
\EQ
where $bb_{i,j}(P,Q) $ is defined in Eq.~\ref{eq: bb1}. We continue with computing the expectation of a pair of points to be BBP, over all possible samples of P and Q, denoted by $E_{\text{BBP}}$. That is,  \vspace{-0.01cm}
{\small \BQ \label{eq:E_bbp}
E_{\text{BBP}} = \iint\limits_{P,Q} \left. bb_{i,j}(P, Q)\Pr\{P\}\Pr\{Q\}dPdQ,\right.
\vspace{-0.07cm}\EQ}
This is a multivariate integral over all points in P and Q. However, assuming each point is independent of the others
this integral can be simplified as follows.
\vspace{-0.1cm}

\minesubsec{Claim:} \vspace{-0.15cm}
\BQ
\label{eq:claim_1d}
\begin{array}{ll}
	E _{\text{BBP}} = &
	\iint\limits_{-\infty}^{~~~\infty} (F_Q(p^-)\! + \! 1 \!- \!F_Q(p^+) )^{N_Q-1}\cdot  \\
&~~~~~(F_P(q^-)\!+\!1\!-\!F_P(q^+) )^{N_P-1}f_P(p)f_Q(q)dpdq,
\end{array}
\EQ
where, $F_P(x)$, and $F_Q(x)$ denote the CDFs of P and Q, respectively. That is, $F_P(x)\!=\!\Pr\{p \leq x\}$. And, $p^- \!=\!p-d(p,q)$, $p^+\!=\!p + d(p,q)$, and $q^{+}, q^-$ are similarly defined.

\minesubsec{Proof:} Due to the independence between the points, the integral in Eq.\ref{eq:E_bbp} can be decoupled as follows:
\BQ \label{eq:E_bbp2}
\begin{array}{l}\
	E_{\text{BBP}} =  \\
	\int\limits_{p_1}\cdots\int\limits_{p_{N_P}} \int\limits_{q_1}\cdots\int\limits_{q_{N_Q}} bb_{i,j}(P, Q)\prod\limits_{k=1}^{N_P}  f_P(p_k) \prod\limits_{l=1}^{N_Q} f_Q(q_l)dPdQ  \\
\end{array}
\EQ
With abuse of notation, we use $dP=dp_1 \cdot dp_2\cdots dp_N$, and $dQ=dq_1 \cdot dq_2 \cdots dq_M$.
Let us consider the function $bb_{i,j}(P, Q)$ for a given realization of P and Q. By definition, this indicator function equals 1 when $p_i$ and $q_j$ are nearest neighbors of each other, and zero otherwise. This can be expressed in terms of the distance between the points as follows:
\BQ \begin{array}{l}
 bb_{i,j}(P, Q)  = \\\hspace{-0.1cm} \prod\limits_{k\neq i, k=1}^{N_P}\mathbb{I}[d(p_k, q_j) > d(p_i, q_j)]\prod\limits_{l\neq j, l=1}^{N_Q}\mathbb{I}[d(q_l, p_i) > d(p_i, q_j)]
\end{array}
\EQ
where $\mathbb{I}$ is an indicator function. It follows that for a given value of $p_i$ and $q_j$, the contribution of $p_k$ to the integral in  Eq.~\ref{eq:E_bbp2} can be decoupled. Specifically, we define:
\BQ Cp_k = \int\limits_{-\infty}^{\infty}  \mathbb{I}[d(p_k, q_j) > d(p_i, q_j)]f_P(p_k)dp_k
\EQ
Assuming $d(p,q)=\sqrt{(p-q)^2} = |p-q|$, the latter can be written as:
{\small\BQ \label{eq: bb_pq2}
		\begin{array}{l}
			Cp_k  = \int\limits_{-\infty}^{\infty}  \mathbb{I}[p_k \!<\! q_j^- \vee p_k \!>\! q_j^+ ]f_P(p_k)dp_k
		\end{array} \vspace{-0.2cm}
		\EQ}
	where  $q_j^- \!=\! q_j\!-\!d(p_i,q_j)$ , $q_j^+ \!=\! q_j\!+\! d(p_i,q_j)$. Since $q_j^- < q_j^+$, it can be easily shown that $Cp_k$ can be expressed in terms of $F_P(x)$, the CDF of P:
{\small	\BQ
Cp_k =  F_P(q_j^-)\! + \! 1 \!- \!F_P(q_j^+) \\
	\EQ}
The same derivation hold for computing  $Cq_l$, the contribution of $q_l$ to the integral in Eq.~\ref{eq:E_bbp2}, given $p_i$, and $q_j$. That is,
	\BQ
	Cq_l =  F_Q(p_i^-)\! + \! 1 \!- \!F_Q(p_i^+) \\
	\EQ
	where $p_i^-, p_i^+$ are similarly defined and $F_Q(x)$ is the CDF of Q. Note that $Cp_k$ and $Cq_l$ depends only on $p_i$ and $q_j$ and on the underlying distributions. Therefore,  Eq.~\ref{eq:E_bbp2} results in:
\BQ
\label{eq:E_BBP}
\begin{array}{rl} E_{\text{BBP}} =\! & \hspace{-0.1cm} \iint\limits_{p_i,q_j}dp_idq_jf_P(p_i)f_Q(q_j) \prod\limits_{\small k=1, k\neq i}^{N_P} \hspace{-0.2cm} Cp_k \prod\limits_{l=1, l\neq j}^{N_Q} \hspace{-0.2cm} Cq_l \\
\vspace{0.2cm}= &  \hspace{-0.1cm}\iint\limits_{p_i,q_j}dp_idq_jf_P(p_i)f_Q(q_j) Cp_k^{N_P-1}Cq_l^{N_Q-1} \\
\end{array}
\EQ
Substituting the expressions for $Cp_k$ and $Cq_l$ in Eq.~\ref{eq:E_BBP}, and omitting the subscripts $i,j$ for simplicity, result in Eq.~\ref{eq:claim_1d}, which completes the proof.

 In general, the integral in Eq.~\ref{eq:claim_1d} does not have a closed form solution, but it can be solved numerically for selected underlying distributions. To this end, we proceed with Gaussian distributions, which are often used as simple statistical models of image patches.
  \comment{\begin{figure}
   	\centering
   	\includegraphics[width = 0.42\textwidth]{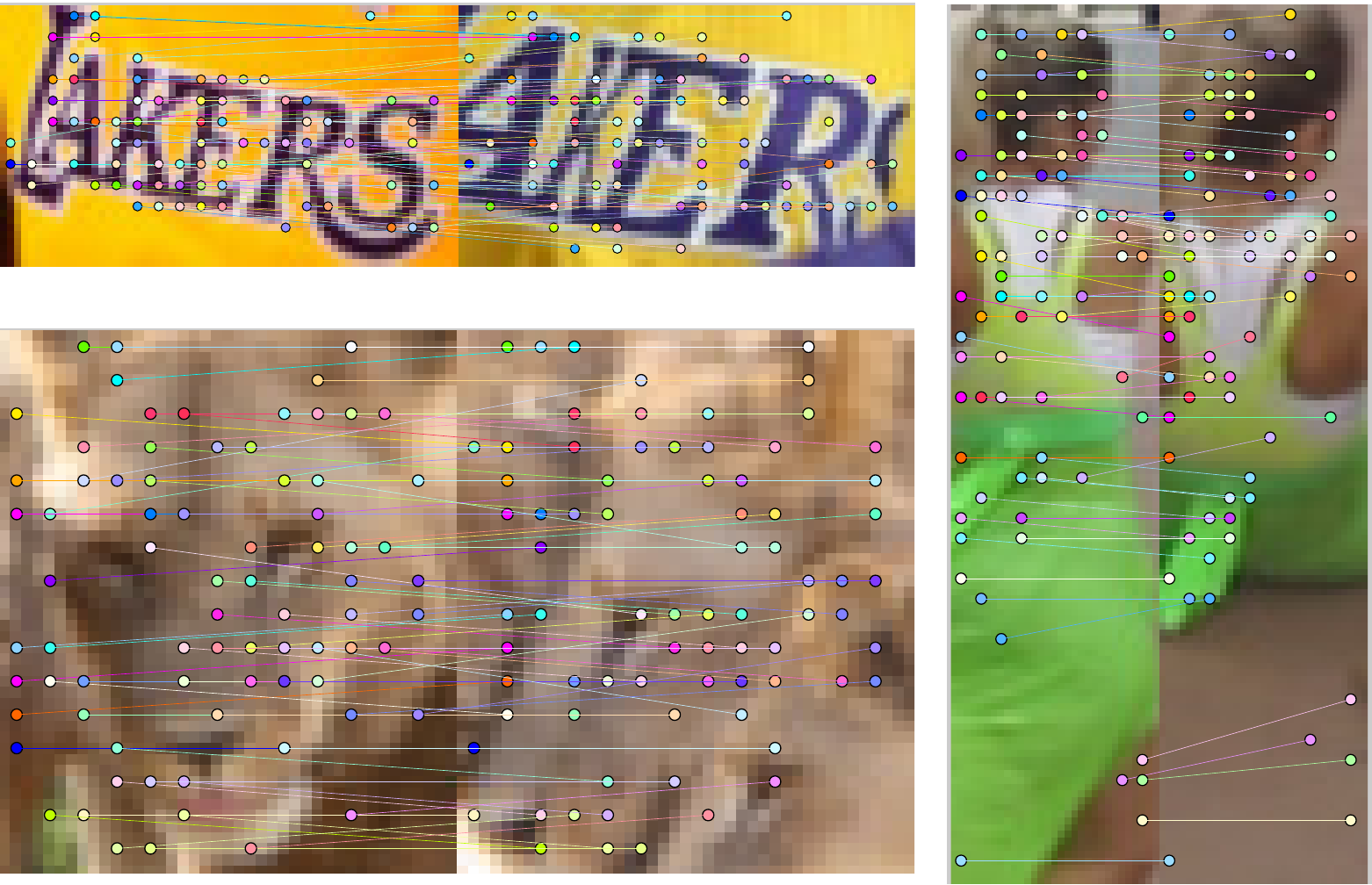}
   	\caption{{\bf Visulization of the BBP:} the BBP between the template and the BBS detection results (see Fig.~\ref{fig: res}). The best buddies are marked in the same color and connected by a line.}
   	\label{fig:BBP_res}
   \end{figure}}
 We then use Monte-Carlo integration to approximate $E_{\text{BBP}}$ for discrete choices of parameters $\mu$ and $\sigma$ of $Q$ in the range of [0, 10] while fixing the distribution of $P$ to have $\mu=0, \sigma=1$. We also fixed the number of points to $N_P=N_Q=100$.
The resulting approximation for $E_{\text{BBP}}$ as a function of the parameters $\mu,\sigma$ is shown in Fig.~\ref{fig:theory}, on the left.  As can be seen, $E_{\text{BBP}}$ is the highest at $\mu=0, \sigma=1$, i.e., when the points are drawn from the same distribution, and drops rapidly as the the underlying distribution of $Q$ deviates from $N(0,1)$.

Note that  $E_{\text{BBP}}$  does not depends on $p$ and $q$ (because of the integration, see Eq.~\ref{eq:claim_1d}. Hence, the expected value of the BBS between the sets (Eq.~\ref{eq:bb_E}) is given by:
\BQ 	E[\text{BBS}(P,Q)]=c\cdot E_{\text{BBP}}\EQ \vspace{-0.01cm}
where $c = \frac{N_PN_Q}{\min\{N_P,N_Q\} } $ is constant.

We can compare the BBS to the expectation of SSD, and SAD. The expectation of the SSD has a closed form solution given by: \vspace{-0.05cm}
{\small\BQ E[\text{SSD(P,Q)}]=\iint\limits_{-\infty}^{~~~~\infty}(p-q)^2f_P(p)f_Q(q|k)dpdq = 1+\mu^2 + \sigma^2.
\EQ}
Replacing $(p-q)^2$ with $|p-q|$ results in the expression of the SAD. In this case, the expected value reduces to the expectation of the Half-Normal distribution and is given by:\vspace{-0.1cm}
{\small \BQ E[\text{SAD(P,Q)}] = \frac{1}{\sqrt{2\pi}} \sigma_K\exp^{-\mu^2/(2\sigma^2)}+\mu(1-2f_P(-\mu/\sigma))
\EQ}
Fig.~\ref{fig:theory}(b)-(c) shows the maps of the expected values for $1-\text{SSD}_n(P,Q)$, and $1-\text{SAD}_n(P,Q)$, where $\text{SSD}_n, \text{SAD}_n$ are the expectation of SSD and SAD, normalized to the range of [0,1]. As can be seen, the SSD and SAD results in a much wider spread around their mode. Thus, we have shown that the likelihood of a pair of points to be a BBP  (and hence the expectation of the BBS) is the highest when  P and Q are drawn from the same distribution and drops sharply as the distance between the distributions increases.  This makes the BBS a robust and distinctive measure that results in well-localized modes.

\minesubsec{Multi-dimensional Case:} 
  With the result of the 1D case in hand, we can bound the expectation of BBS when $P$ and $Q$ are sets of multi-dimensional points, i.e., $p_i,q_j \in R^d$.

If the $d$-dimensions are uncorrelated (i.e., the covariance matrices are diagonals in the Gaussian case),  a sufficient (but not necessary) condition for a pair of points to be BBP is that the point would be BBP in each of the dimensions. In this case, the analysis can be done for each dimension independently similar to what was done in Eq.~\ref{eq:claim_1d}. The expectation of the BBS in the multi-dimensional case is then bounded by the product of the expectations in each of the dimensions. That is,
\vspace{-0.2cm}
\BQ
E_{\text{BBS}} \geq \prod\limits_{i=1}^d E^i_{\text{BBS}},
\EQ
where  $E^i_{\text{BBS}}$ denote the expectation of BBS in the $i^\text{th}$ dimension. This means that the BBS is expected to be more distinctive, i.e., to drop faster as $d$ increases. Note that if a pair of points is not a BBP in one of the dimensions, it does not necessarily imply that the multi-dimentional pair is not BBP. Thus, this condition is sufficient but not necessary.

\subsection{BBS and Chi-Square}
Chi-Square is often used to measure the distance between histograms of two sets of features. For example, in face recognition, $\chi^2$ is used to measure the similarity between local binary patterns (LBP) of two faces~\cite{ahonen2006face}, and it achieves superior performance relative to other distance measures. 

In this section, we will discuss the connection between this well known statistical distance measure and BBS. Showing that, for sufficiently large point sets, BBS converges to the $\chi^2$ distance.

\begin{figure}
\centering
\begin{tabular}{cc}
\includegraphics[width = 0.32\columnwidth]{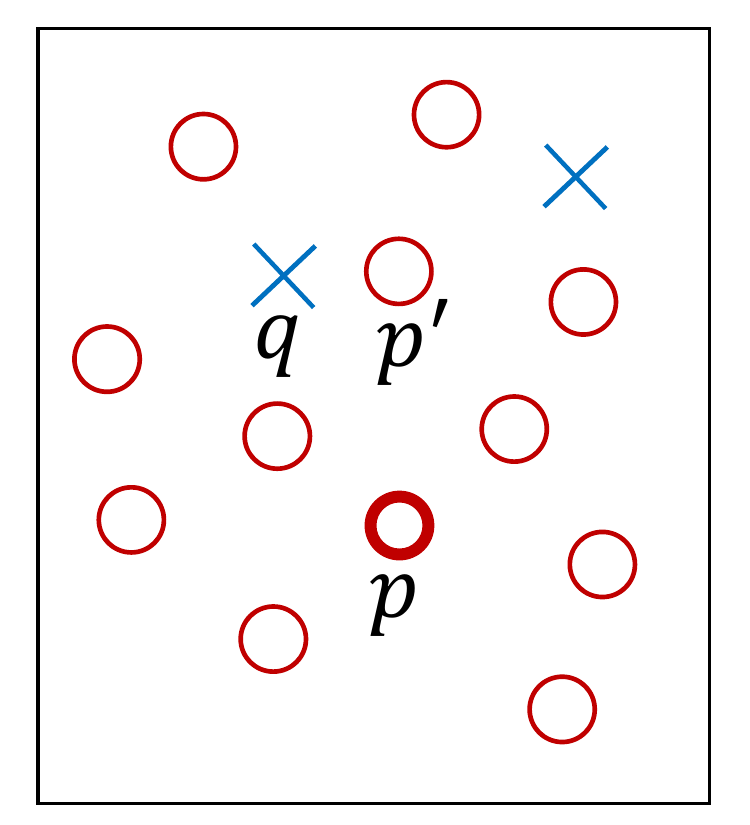} &
\includegraphics[width = 0.32\columnwidth]{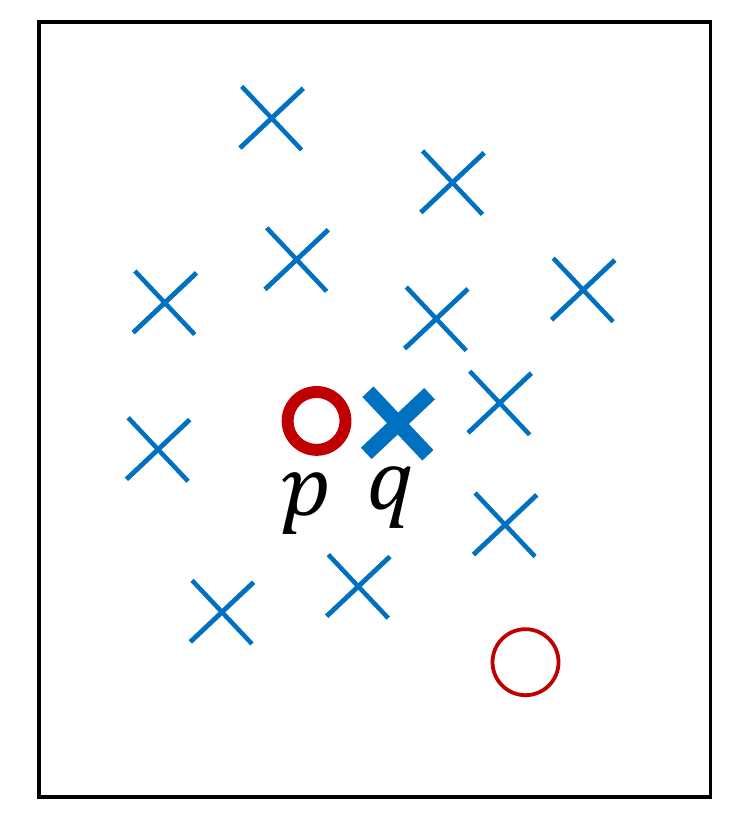} \\
(a) & (b)\\
\end{tabular}
\caption{{\bf Finding a Best-Buddy:} We illustrate how the underlying density functions affect the probability that a point $p$ (bold red circle) has a best buddy.
(a) Points from set $P$ (red circles) are dense but points from set $Q$ (blue cross) are sparse.
Although $q$ is the nearest neighbor of $p$ in $Q$, $p$ is not the nearest neighbor of $q$ in $P$ ($p'$ is closer).
(b) Points from set $Q$ are dense and points from set $P$ are sparse.
In this case, $p$ and $q$ are best buddies, as $p$ is the closest point to $q$.}
\label{fig:bestbuddy_p}
\end{figure}

We assume, as before, that point sets $P$ and $Q$ are drawn i.i.d. from 1D distribution functions $f_P(p)$ and $f_Q(q)$ respectively. We begin by considering the following lemma:
\begin{lemma}\label{lemma:1}
Given a point $p_i=p$ in $P$, let $Pr[bb(p_i=p;P,Q)]$ be the probability that $p_i$ has a best buddy in $Q$. Then we have:
\BQ
\limitN Pr[bb(p_i=p;P,Q)] = \frac{f_Q(p)}{f_P(p) + f_Q(p)},\label{eq:claim}
\EQ
\end{lemma}
For the proof of the lemma see appendix~\ref{A:lemma_proof}. Intuitively, if there are many points from $P$ in the vicinity of point $p$, but only few points from $Q$, i.e. $f_P(p)$ is large but $f_Q(p)$ is small. It is then hard to find a best buddy in $Q$ for $p$, as illustrated in Figure~\ref{fig:bestbuddy_p}(a). Conversely, if there are few points from $P$ in the vicinity of $p$ but many points from $Q$, i.e. $f_P(p)$ is small and $f_Q(p)$ is large. In that case, it is easy for $p$ to find a best buddy, as illustrated in Figure~\ref{fig:bestbuddy_p}(b).

\begin{figure*}
\centering
\begin{tabular}{cc}
\includegraphics[width = 0.45\textwidth]{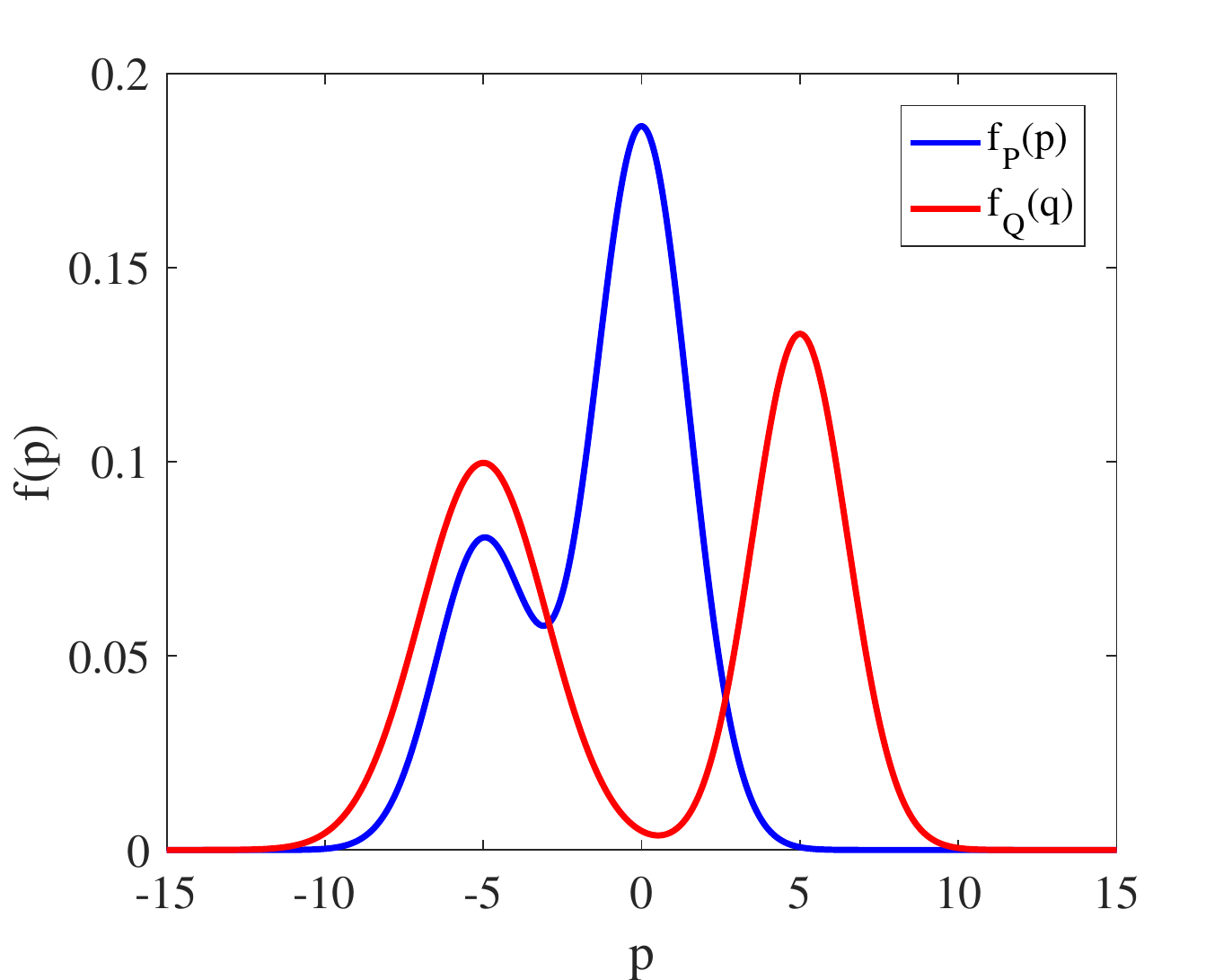} &
\includegraphics[width = 0.45\textwidth]{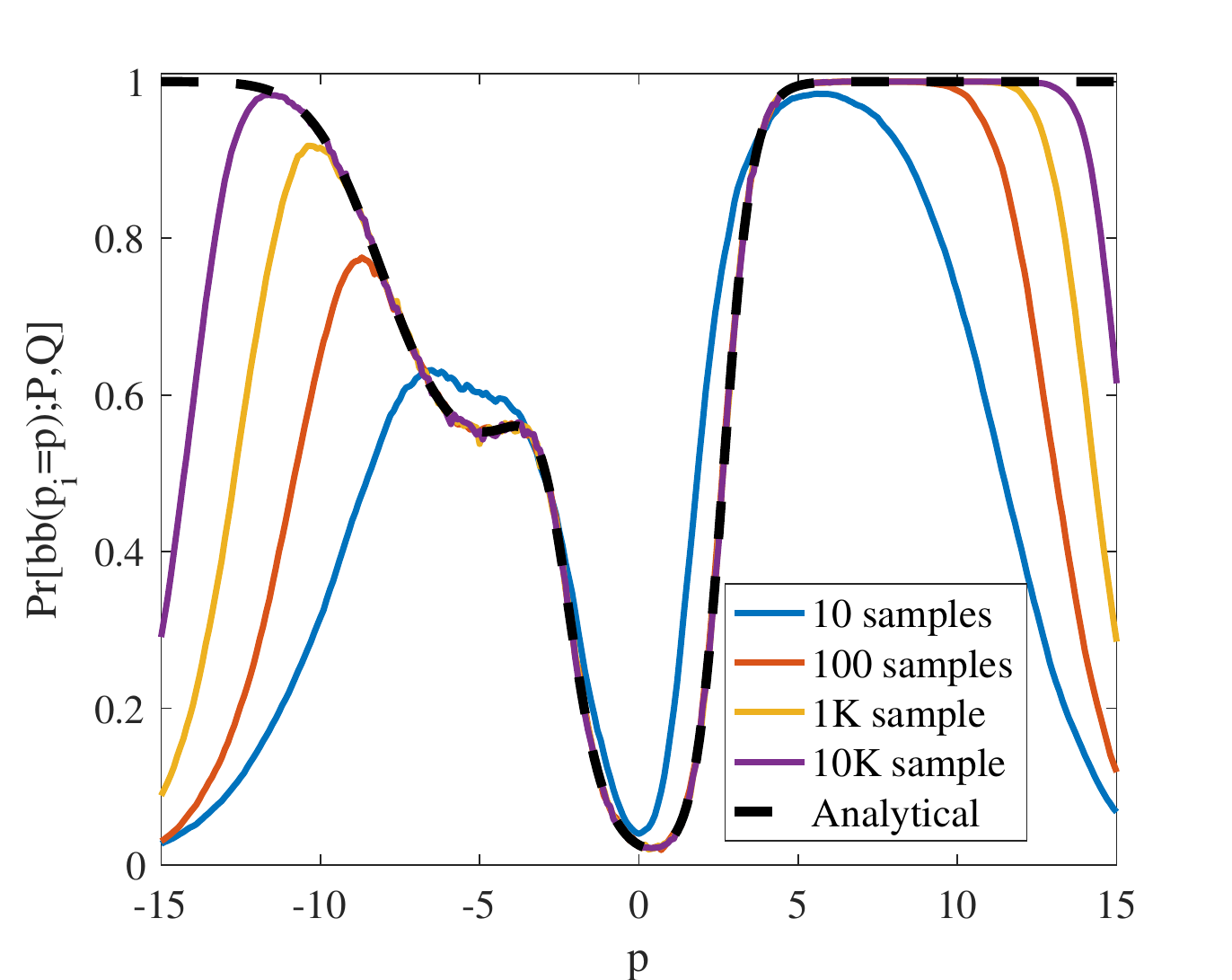} \\
(a) Density functions of two Gaussian mixtures & (b) The probability that a point in P has a best buddy
\end{tabular}
\caption{{\bf Illustrating Lemma~\ref{lemma:1}:}
Point sets $P$ and $Q$ are sampled iid from the two Gaussian mixtures shown in (a). The probability that a point in set $P$ has a best buddy in set $Q$ is empirically computed for different set sizes (b). When the size of the sets increase, the empirical probability converges to the analytical solution in Lemma~\ref{lemma:1} (dashed black line).}
\label{fig:lamma1_sim}
\end{figure*}

A synthetic experiment illustrating the lemma is shown in Figure~\ref{fig:lamma1_sim}. Two Gaussian mixtures, each consisting of two 1D-Gaussian distributions are used (Figure~\ref{fig:lamma1_sim}(a)). Sets $P$ and $Q$ are sampled from these distributions (each set from a different mixture model). We then empirically calculate the probability that a certain point $p_i$ from set $P$ has a best buddy in set $Q$ for different set sizes, ranging from 10 to 10000 points, Figure~\ref{fig:lamma1_sim}(b) .
As the sets size increases, the empirical probability converges to the analytical value given by Lemma~\ref{lemma:1}, marked by the dashed black line.
Note how the results agree with our intuition. For example, at $p=0$, $f_P(p)$ is very large but $f_Q(p)$ is almost $0$, such that $Pr[bbs(p_i;P,Q)]$ is almost 0. At $p=5$, however, $f_P(p)$ is very small and $f_Q(p)$ is almost $0$, so $Pr[bbs(p_i;P,Q)]$ is almost 1.

Lemma~\ref{lemma:1} assumes the value of the point $p_i$ is fixed. However, we need to consider that $p_i$ itself is also sampled from the distribution $f_P(p)$, in which case the probability this point has a best buddy is:
\BQ
\begin{array}{l}
Pr[bb(p_i;P,Q)] = \\ \int\limits_{p=-M}^M f_P(p) \cdot Pr(p_i=p;P,Q) dp.
						   = \int\limits_{p=-M}^M \frac{f_Q(p)f_P(p)}{f_P(p) + f_Q(p)} dp
						   \end{array}
\EQ
Where we assume both density functions are defined on the closed interval $[-M,M]$.

We are now ready to show that BBS converges to Chi-Square,
\begin{theorem} \label{theorem:1}
Suppose both density functions are defined on a close interval $[-M,M]$, non-zero and Lipschitz continuous
\footnote{Note that most of density functions, like the density function of a Gaussian distribution, are non-zero and Lipschitz continuous in their domain.}.
 That is,
\begin{enumerate}
\item $\forall p,q$, $f_P(p) \neq 0, f_Q(q) \neq 0$ \vspace{0.2cm}
\item $\exists A>0,\forall p,q,h,~s.t.~|f_P(p+h)-f_P(p)| < A|h|$ and $|f_Q(q+h)-f_Q(q)| < A|h|$,
\end{enumerate}
then we have,
\BQ
\begin{array}{lll}
\limitN E[\emph{BBS}(P,Q)] &=&  \intp \frac{f_P(p)f_Q(p)}{f_P(p) + f_Q(p)} d p\\ \\ &=& \frac{1}{2} - \frac{1}{4}\chi^2(f_p, f_q),
\end{array}
\label{eq:theorem1}
\EQ
\end{theorem}
\noindent where $\chi^2(f_p, f_q)$ is the Chi-Square distance between two distributions.  

To see why this theorem holds, consider the BBS measure between two sets, $P$ and $Q$. When the two sets have the same size, the BBS measure equals to the fraction of points in $P$ that have a best buddy, that is $BBS(P,Q) = \frac{1}{N}\sum_{i=1}^N bbs(p_i;P,Q)$. Taking expectation on both sides of the equation, we get:
\BQ
\begin{array}{lll}
 E[BBS(P,Q)]   &=& \frac{1}{N}\sum_{i=1}^N E[bbs(p_i;P,Q)] \vspace{0.3cm}\\
 &= & \frac{1}{N} \cdot N \cdot E[bbs(p_i;P,Q)] \vspace{0.3cm}\\
 &= & \int\limits_{p=-M}^M \frac{f_Q(p)f_P(p)}{f_P(p) + f_Q(p)} dp.
\end{array}
\EQ
Where for the last equality we used lemma~\ref{lemma:1}. This completes the proof of Theorem~\ref{theorem:1}.

The theorem helps illustrate why BBS is robust to outliers. To see this, consider the signals in Figure ~\ref{fig:lamma1_sim}(a). As can be seen $f_P$ and $f_Q$ are both Gaussian mixtures. Let us assume that the Gaussian with mean $-5$ represents the foreground (in both signals), i.e. $\mu_{fg}=-5$, and that the second Gaussian in each mixture represents the background, i.e. $\mu_{bg1} = 0$ and $\mu_{bg2} = 5$. Note how, $f_P(p)$ is very close to zero around $\mu_{bg2}$ and similarly $f_Q(q)$ is very close to zero around $\mu_{bg1}$. This means that the background distributions will make very little contribution to the $\chi^2$ distance, as the numerator $f_P(p)f_Q(q)$ of Eq.~\ref{eq:theorem1} is very close to 0 in both cases.

We note that using BBS has several advantages compared to using $\chi^2$ . One such advantage is that BBS does not require binning data into histograms. It is not trivial to set the bin size, as it depends on the distribution of the features. A second advantage is the ability to use high dimensional feature spaces. The computational complexity and amount of data needed for generating histograms quickly explodes when the feature dimension goes higher. On the contrary, the nearest neighbor algorithm used by BBS can easily scale to high-dimensional features, like Deep features.
	\section{Implementation Details}
In this section we provide information on the specific feature spaces used in our experiments. Additionally, we analyze the computational complexity of BBS and propose a caching scheme allowing for more efficient computation.

\subsection{Feature Spaces} \label{sec:details}
In order to perform template matching BBS is computed, exhaustively, in a sliding window. A joint spatial-appearance representation is used in order to convert both template and candidate windows into point sets. For the spatial component normalized $xy$ coordinates within the windows are used. For the appearance descriptor we experiment with both color features as well as Deep features.

\minesubsec{Color features:} 
When using color features, we break the template and candidate windows into $k \times k$ distinct patches. Each such $k \times k$ patch is represented by its $3\cdot k^2$ color channel values and $xy$ location of the central pixel, relative to the patch coordinate system.
For our toy examples and qualitative experiments $RGB$ color space is used. However, for our  quantitative evaluation $HSV$ was used as it was found to produced better results. Both spatial and appearance channels were normalized to the range $[0,1]$  .
The point-wise distance measure used with our color features is:
\BQ \label{eq:dist color} \textsl{\textsl{}}
d(p_i,q_j) = ||p_i^{(A)}-q_j^{(A)}||^2_2+\lambda||p_i^{(L)}-q_j^{(L)}||^2_2
\EQ
where superscript $A$ denotes a points appearance and superscript $L$ denotes a points location. The parameter $\lambda=0.25$ was chosen empirically and was fixed in all of our experiments.\\

\minesubsec{Deep features:} 
For our Deep feature we use features taken from the VGG-Deep-Net~\cite{Simonyan14c}. Specifically, we take features from two layers of the network,  conv$1\_2$ (64 features) and conv$3\_4$ (256 features). The feature maps from conv$1\_2$ are down-sampled twice, using max-pooling, to reach the size of the conv$3\_4$ which is down-sampled by a factor of $1/4$ with respect to the original image. In this case we treat every pixel in the down-sampled feature maps as a point. Each such point is represented by its $xy$ location in the down-sampled window and its appearance is given by the 320 feature channels. Prior to computing the point-wise distances each feature channel is independently normalized to have zero mean and unit variance over the window.
The point-wise distance in this case is:
\BQ \label{eq:dist } \textsl{\textsl{}}
d(p_i,q_j) = <p_i^{(A)},q_j^{(A)}> + exp(-\lambda||p_i^{(L)}-q_j^{(L)}||^2_2)
\EQ
where $<\cdot,\cdot>$ denotes the inner product operator between feature vectors. Unlike the color features we now want to maximize $d$ rather then minimize it (we can always minimize $-d$). The parameter $\lambda=1$ was chosen empirically and was fixed in all of our experiments.

\begin{figure*}[t!] 
  	\centering
		\includegraphics[width = 1.0\textwidth]{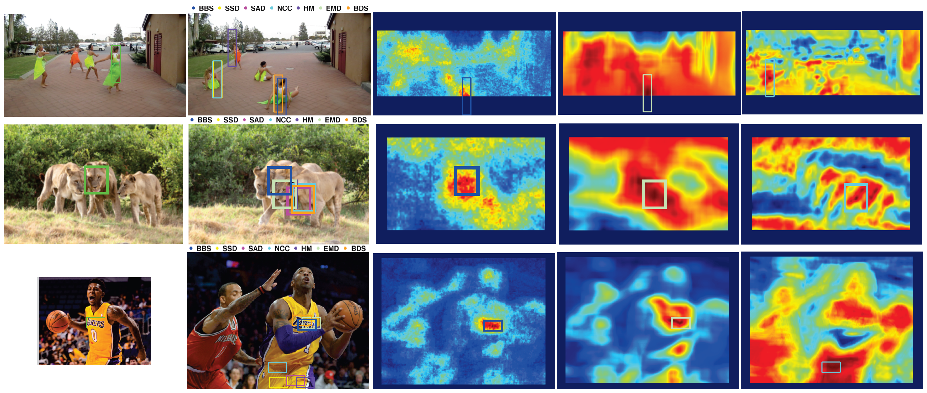} \\
  	\caption{{\bf BBS results on Real Data:} (a), the templates are marked in green over the input images. (b) the target images marked with the detection results of 6 different methods (see text for more details). BBS results are marked in blue. (c)-(e), the resulting likelihood maps using BBS, EMD and NCC , respectively; each map is marked with the detection result, i.e., its global maxima.}
  	\label{fig: res}
  \end{figure*}

\subsection{Complexity}
Computing BBS between two point sets $P,Q\in\mathbb{R}^d$, requires computing the distance between each pair of points. That is, constructing a distance matrix $D$ where $[D]_{i,j}\!=\!d(p_i,q_j)$. Given $D$, the nearest neighbor of $p_i\in P$, i.e. $NN(p_i,Q)$, is the minimal element in the $i^{th}$ row of $D$. Similarly, $NN(q_j,P)$ is the minimal element in the $j^{th}$ column of $D$. BBS is then computed by counting the number of mutual nearest neighbors (divided by a constant).\\
In this section we analyze the computational complexity of computing BBS exhaustively for every window in a query image. We then propose a caching scheme, allowing extensive computation reuse which dramatically reduces the computational complexity, trading it off with increased memory complexity. 

\minesubsec{Naive implementation:}  
For our analysis we consider a target window of size $w\times h$ and some query image $I$ of size $W\times H$. Both represented using a feature space with $d$ feature channels. Let us begin by considering each pixel in our target window as a point in our target point set $P$ and similarly every pixel in some query window is considered as a point in the query point set $Q$. In this case, $|P|=|Q| = w\cdot h \triangleq l$ and our distance matrices $D$ are of size $l \times l$. Assuming some arbitrary image padding, we have $W\cdot H \triangleq L$ query windows for which BBS has to be computed. Computing all the $L$ distance matrices requires $O(Ll^2d)$. For each such distance matrix we need to find the minimal element in every row and column. The minimum computation for a single row or column is done in $O(l)$ and for the entire matrix in $O(l^2)$. Therefore, the complexity of computing BBS naively for all query windows of image $I$ is,
\begin{equation}
	O(Ll^4d)
\end{equation}
This is a high computational load compared to simpler methods such as sum-of-square-difference (SSD) that require only $O(Lld)$.

\minesubsec{Distance computation reuse:} When carefully examining the naive scheme above we notice that many pairwise distance computations are performed multiple times. This observation is key to our proposed caching scheme.

Assuming our sliding window works column by column, the first distance matrix in the image has to be fully computed. The second distance matrix, after sliding the query window down by one pixel, has many overlapping distance computations with the previously computed matrix. Specifically, we only have to compute the distances between pixels in the new row added to the query window and the target window. This means we have to recompute only $w$ columns of $D$ and not the entire matrix. Taking this one step further, if we cache all the distance matrices computed along the first image column, then staring from the second matrix in the second column, we would only have to compute the distance  between \textit{one} new candidate pixel and the target window, which means we only have to recompute one column of $D$ which requires only $O(l)$. Assuming $W,H >> w,h$ the majority of distance matrices can be computed in $O(l)$, instead of $O(l^2)$. This means that computing BBS for the entire image $I$ would now require:
\begin{equation}
	O(Ll^3f)
\end{equation}

\minesubsec{Minimum operator load reduction:} So far we have shown how caching can be used to reduce the load of building the distance matrices. We now show how additional caching can reduce the computational load of the minimum operator applied to each row and column of $D$ in order to find the BBP.

As discussed earlier, for the majority of query windows we only have to recompute $one$ column of $D$. This means that for all other $l-1$ columns we have already computed the minimum. Therefore, we actually obtain the minimum over all columns in just $O(l)$.   
For the minimum computation along the rows there are two cases to consider. First, that the minimal value, for a certain row, was in the column that was pushed out of $D$. In this case we would have to find the minimum value for that row, which would require $O(l)$. The second option is that the minimal value of the row was not pushed out and we know where it is from previous computations. In such a case we only have to compare the new element added to the row (by the new column introduced into $D$) relative to the previous minimum value, this operation requires $O(1)$. Assuming the position of the minimal value along a row is uniformly distributed, on average, there will be only one row where the minimum value needs to be recomputed. To see this consider a set of random variables $\{X_i\}_{i=1}^l$ such that $X_i=1$ if and only if the minimal value in the $i$'th row of $D$ was pushed out of the matrix when a new column was introduced. Assuming a uniform distribution $X_i\sim Bernoulli(1/l)$. The number of rows for which the minimum has to be recomputed is given by $m = \sum_{i=1}^{l} X_i$, and the expected number of such rows is,
\begin{equation}
	E[m] = E\left[\sum_{i=1}^{l} X_i\right] = \sum_{i=1}^{l}E[X_i] = \sum_{i=1}^{l}\frac{1}{l}=1
\end{equation}
This means, that on average, there will be only one row for which the minimum has to be computed in $O(l)$ time (for other rows only $O(1)$ is required). Therefore, on average, we are able to find the minimum of all rows and columns in $D$, in $O(l)$ instead on $O(l^2)$.
By combining the efficient minimum computation scheme, along with the reuse of distance computations for building $D$, we reduce the overall BBS complexity over the entire image to,  
\begin{equation}\label{eq:complex3}
	O(Ll^2d)
\end{equation}

\minesubsec{Additional load reduction:} When using color features, we note that the actual complexity of computing BBS for the entire image $I$ is even lower due to the use of non-overlapping $k\times k$ patches instead of individual pixels. This means that both image and target windows are sampled on a grid with spacing $k$ which in turn leads to an overall complexity of:
\begin{equation}
	O\left(\frac{Ll^2d}{k^4}\right)
\end{equation}
We note that the reuse schemes presented above cannot be used with our Deep features due to the fact that we normalize the features differently, with respect to each query window. Also the above analysis does not consider the complexity of extracting the Deep features themselves. 
	\section{Results}
We perform qualitative as well as extensive quantitative evaluation of our method on real world data. We compare BBS with several measures commonly used for template matching. 1) Sum-of-Square-Difference (SSD), 2) Sum-of-Absolute-Difference (SAD), 3) Normalized-Cross-Correlation (NCC), 4) color Histogram Matching (HM) using the $\chi^2$ distance, 5) Bidirectional Similarity~\cite{Simakov2008summarizing} (BDS) computed in the same appearance-location space as BBS. 

\begin{figure*}
 	\centering
 	\begin{tabular}{cc}
 		\includegraphics[width = 0.48\textwidth]{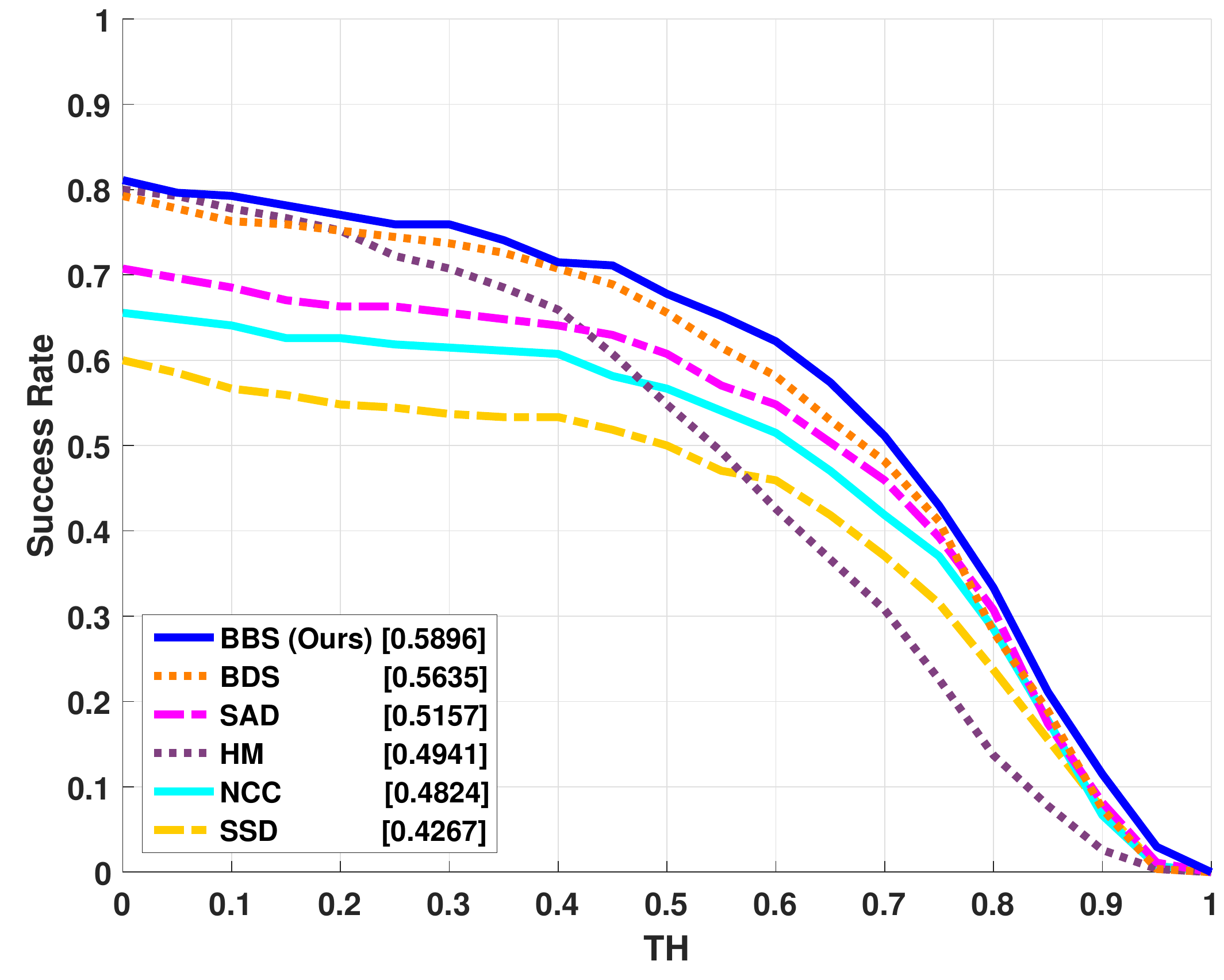} &
		\includegraphics[width = 0.48\textwidth]{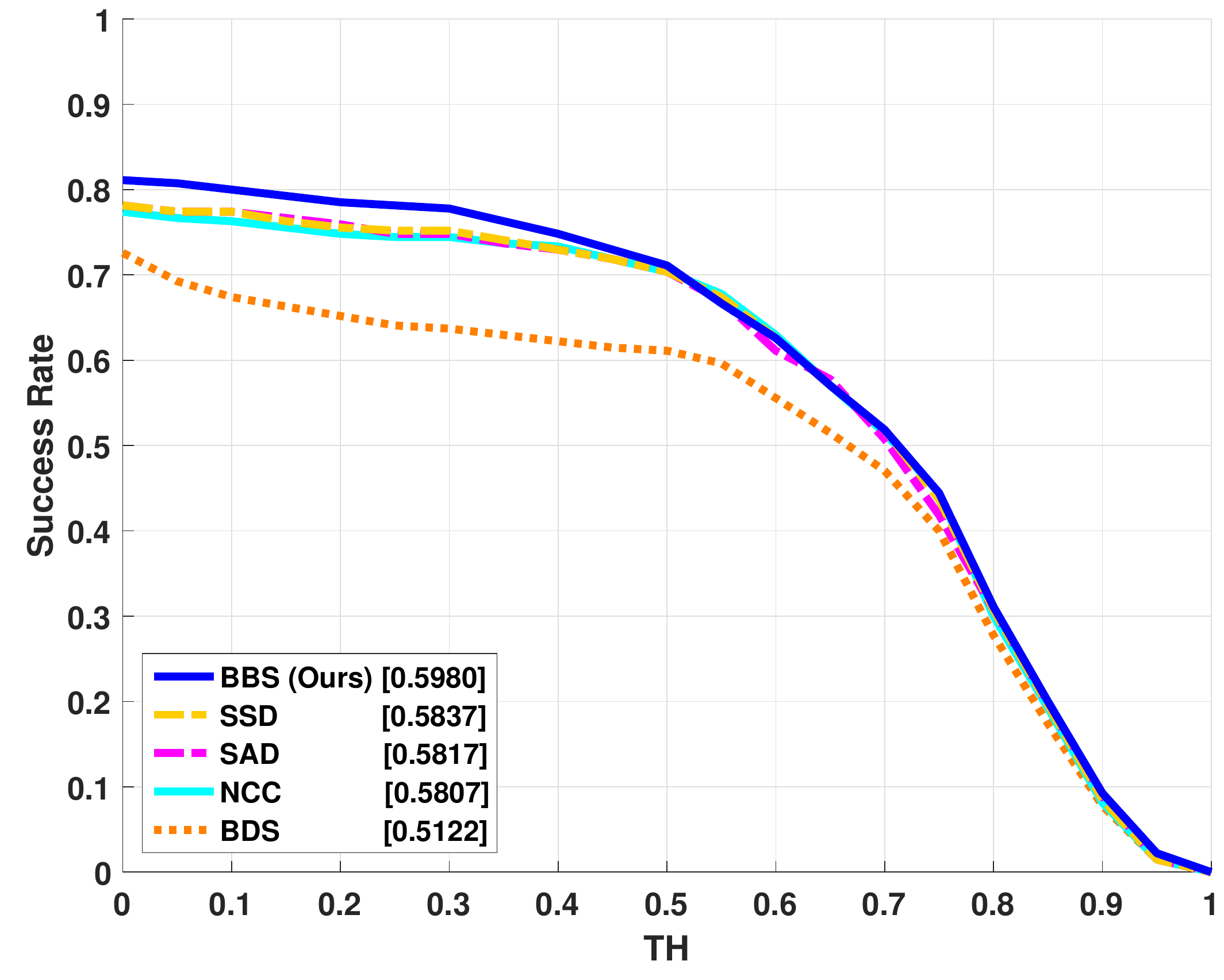} \\
		(a) Color feature, best mode only. & (b) Deep feature, best mode only. \\
		\includegraphics[width = 0.48\textwidth]{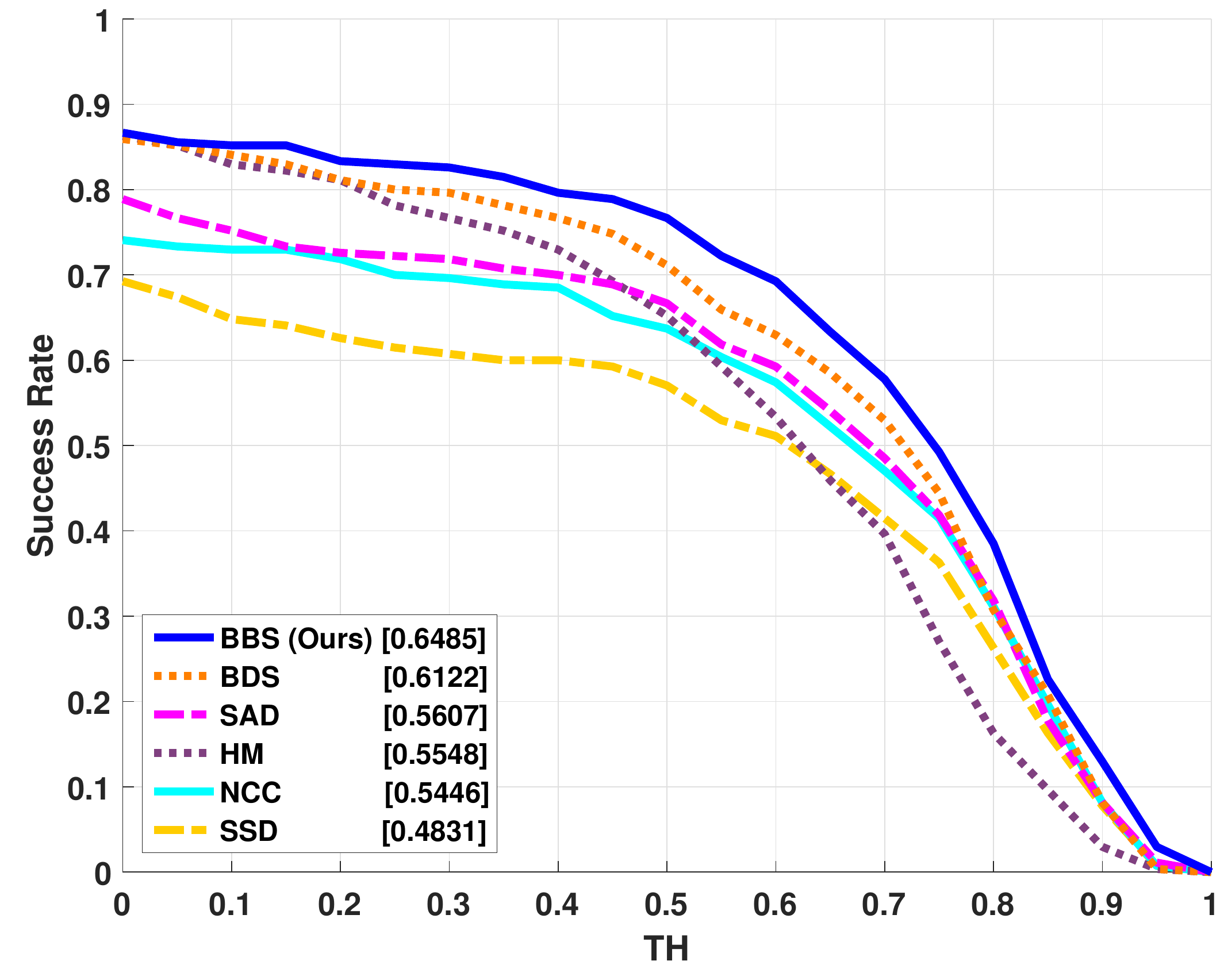} &
		\includegraphics[width = 0.48\textwidth]{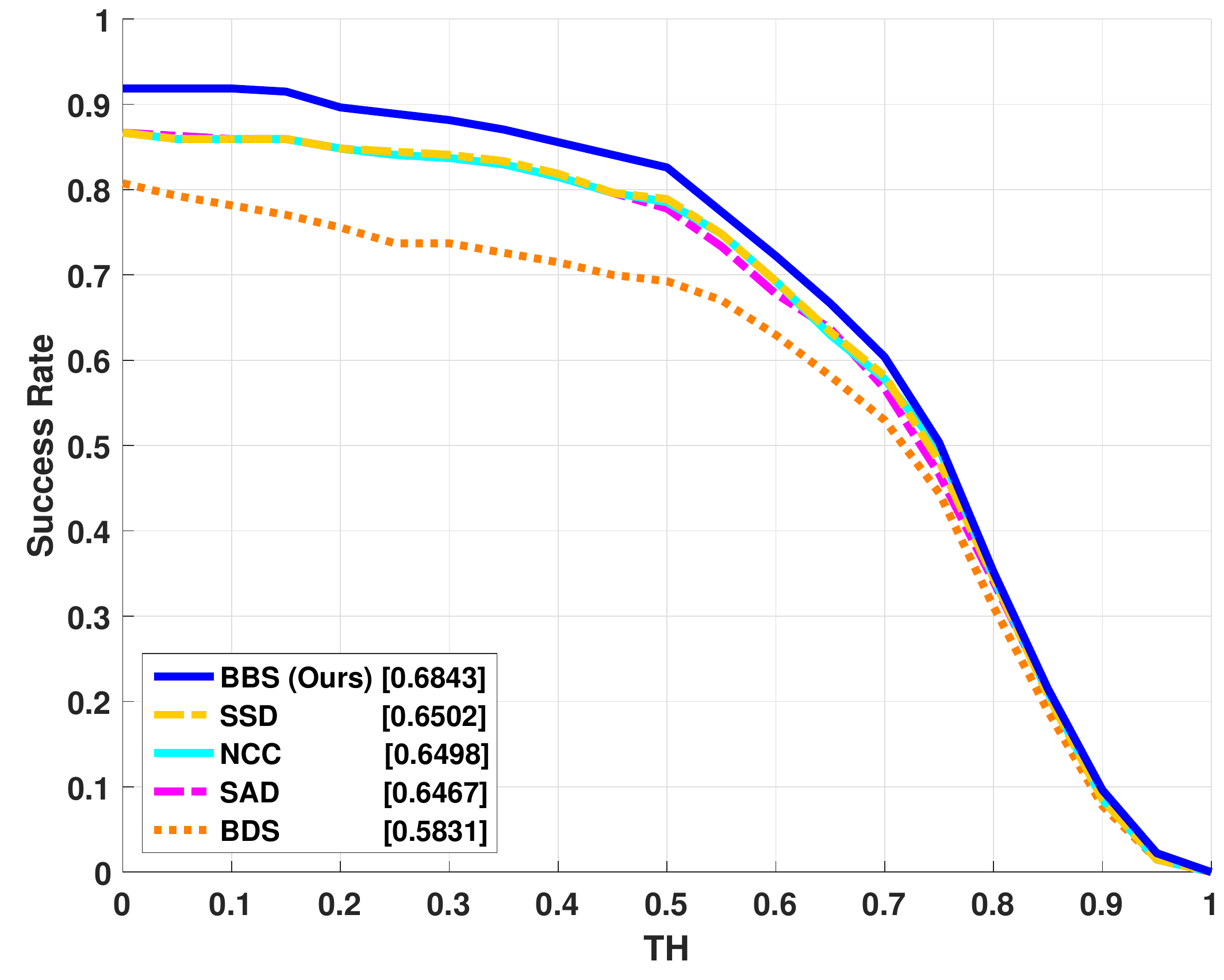} \\
		(c) Color feature, top 3 modes. & (d) Deep feature, top 3 modes. \\		
 	\end{tabular}
 	\caption{{\bf Template matching accuracy:} Evaluation of method performance using 270 template-image pairs with $df=25$. BBS outperforms competing methods as can be seen in ROC curves showing fraction of examples with overlap greater than threshold values in [0,1]. Top: only best mode is considered. Bottom: best out of top 3 modes is taken. Left: Color features. Right: Deep features. Mean-average-precision (mAP) values taken as area-under-curve are shown in the legend. Best viewed in color.}
 	\label{fig: acc} \vspace{-0.3cm}
 \end{figure*}

\begin{figure*}[t!]
 	\centering
 	\begin{tabular}{cccc} 		
		\includegraphics[width = 0.23\textwidth]{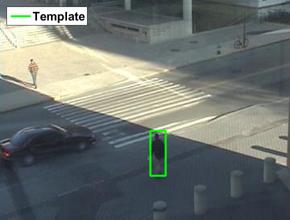} &
		\includegraphics[width = 0.23\textwidth]{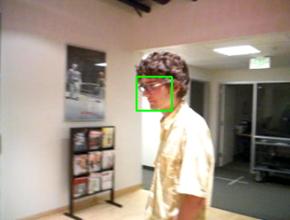} &
		\includegraphics[width = 0.23\textwidth]{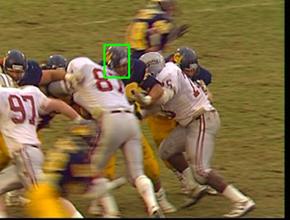} &
		\includegraphics[width = 0.23\textwidth]{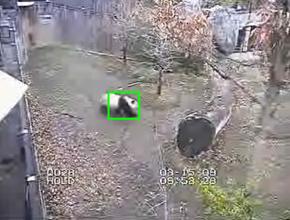} \\
		\includegraphics[width = 0.23\textwidth]{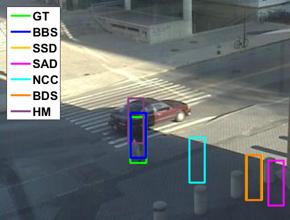} &
		\includegraphics[width = 0.23\textwidth]{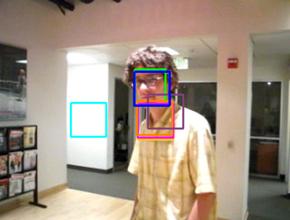} &
		\includegraphics[width = 0.23\textwidth]{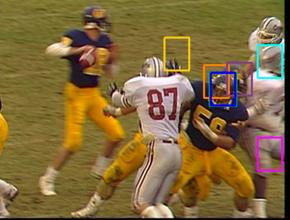} &
		\includegraphics[width = 0.23\textwidth]{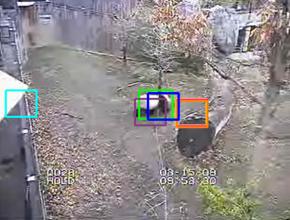} \\
		\includegraphics[width = 0.23\textwidth]{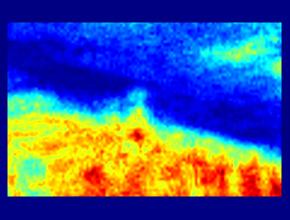} &
		\includegraphics[width = 0.23\textwidth]{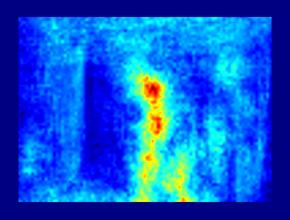} &
		\includegraphics[width = 0.23\textwidth]{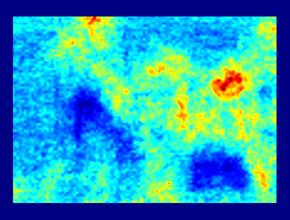} &
		\includegraphics[width = 0.23\textwidth]{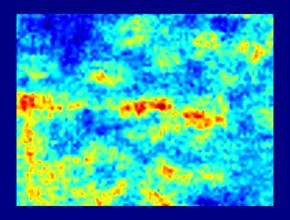} \\
		(a) & (b) & (c) & (d)\\
				
 	\end{tabular}
 	\caption{{\bf Example results using color features}. Top, input images with annotated template marked in green.  Middle,  target images and detected bounding boxes (see legend); ground-truth (GT) marked in green (our results in blue). Bottom, BBS likelihood maps. BBS successfully match the template in all these examples.}
 	\label{fig: examples} 
 \end{figure*}

\begin{figure*}[t!]
 	\centering
 	\begin{tabular}{cccc} 		
		\includegraphics[width = 0.23\textwidth]{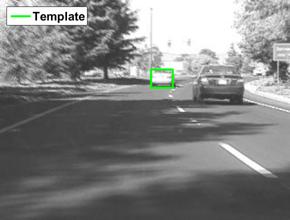} &
		\includegraphics[width = 0.23\textwidth]{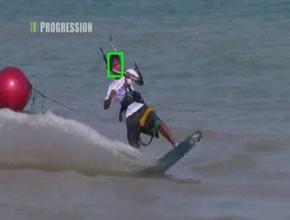} &
		\includegraphics[width = 0.23\textwidth]{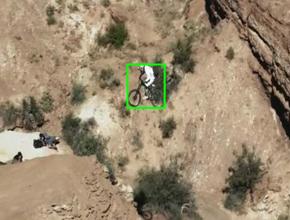} &
		\includegraphics[width = 0.23\textwidth]{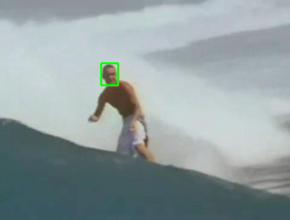} \\
		\includegraphics[width = 0.23\textwidth]{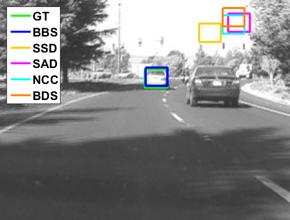} &
		\includegraphics[width = 0.23\textwidth]{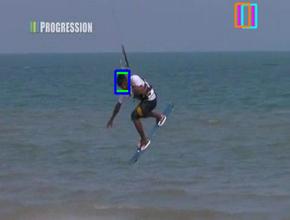} &
		\includegraphics[width = 0.23\textwidth]{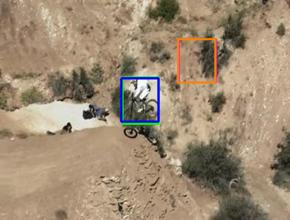} &
		\includegraphics[width = 0.23\textwidth]{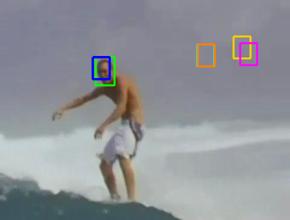} \\
		\includegraphics[width = 0.23\textwidth]{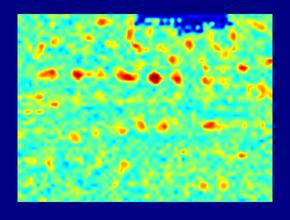} &
		\includegraphics[width = 0.23\textwidth]{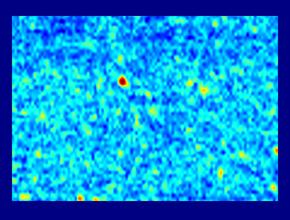} &
		\includegraphics[width = 0.23\textwidth]{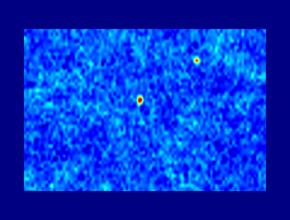} &
		\includegraphics[width = 0.23\textwidth]{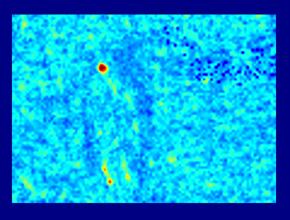} \\
		(a) & (b) & (c) & (d)\\
				
 	\end{tabular}
 	\caption{{\bf Example results using Deep features}. Top, input images with annotated template marked in green.  Middle,  target images and detected bounding boxes (see legend); ground-truth (GT) marked in green (our results in blue). Bottom, BBS likelihood maps. BBS successfully match the template in all these examples.}
 	\label{fig: examples_df} 
 \end{figure*}

\subsection{Qualitative Evaluation} \label{sec:res_qual}
Four template-image pairs taken from the Web are used for qualitative evaluation. The templates, which were manually chosen, and the target images are shown in Figure ~\ref{fig:teaser}(a)-(b), and in Figure ~\ref{fig: res}. In all examples, the template drastically changes its appearance due to large geometric deformation, partial occlusions, and change of background.  

Detection results, using color features with $RGB$ color space, are presented in Figure ~\ref{fig:teaser}(a)-(b), and in Figure ~\ref{fig: res}(b), and compared to the above mentioned methods as well as to the Earth Movers Distance\cite{Rubner00} (EMD). The BBS is the only method successfully matching the template in all these challenging examples. The confidence maps of BBS, presented in Figure ~\ref{fig: res}(c), show distinct and well-localized modes compared to other methods\footnote{Our data and code are publicly available at: \url{http://people.csail.mit.edu/talidekel/Best-Buddies Similarity.html}}. The BBPs for the first example are shown in Figure ~\ref{fig:teaser}(c). As discussed in Sec.~\ref{sec:method}, BBS captures the bidirectional inliers, which are mostly found on the object of interest. Note that the BBPs, as discussed, are not necessarily true physical corresponding points.

\begin{figure*}
 	\centering
 	\begin{tabular}{cc}
 		\includegraphics[width = 0.48\textwidth]{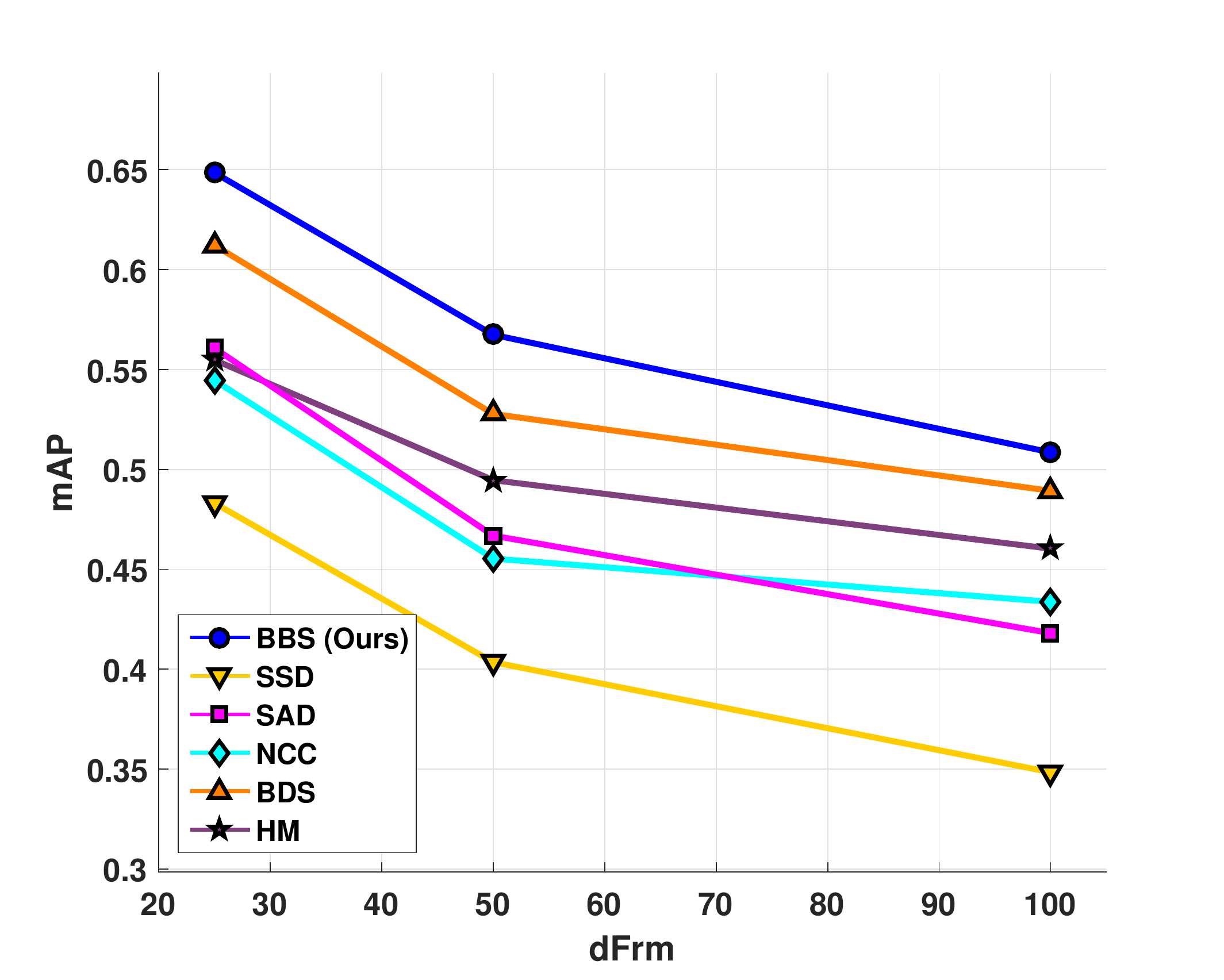} &
		\includegraphics[width = 0.48\textwidth]{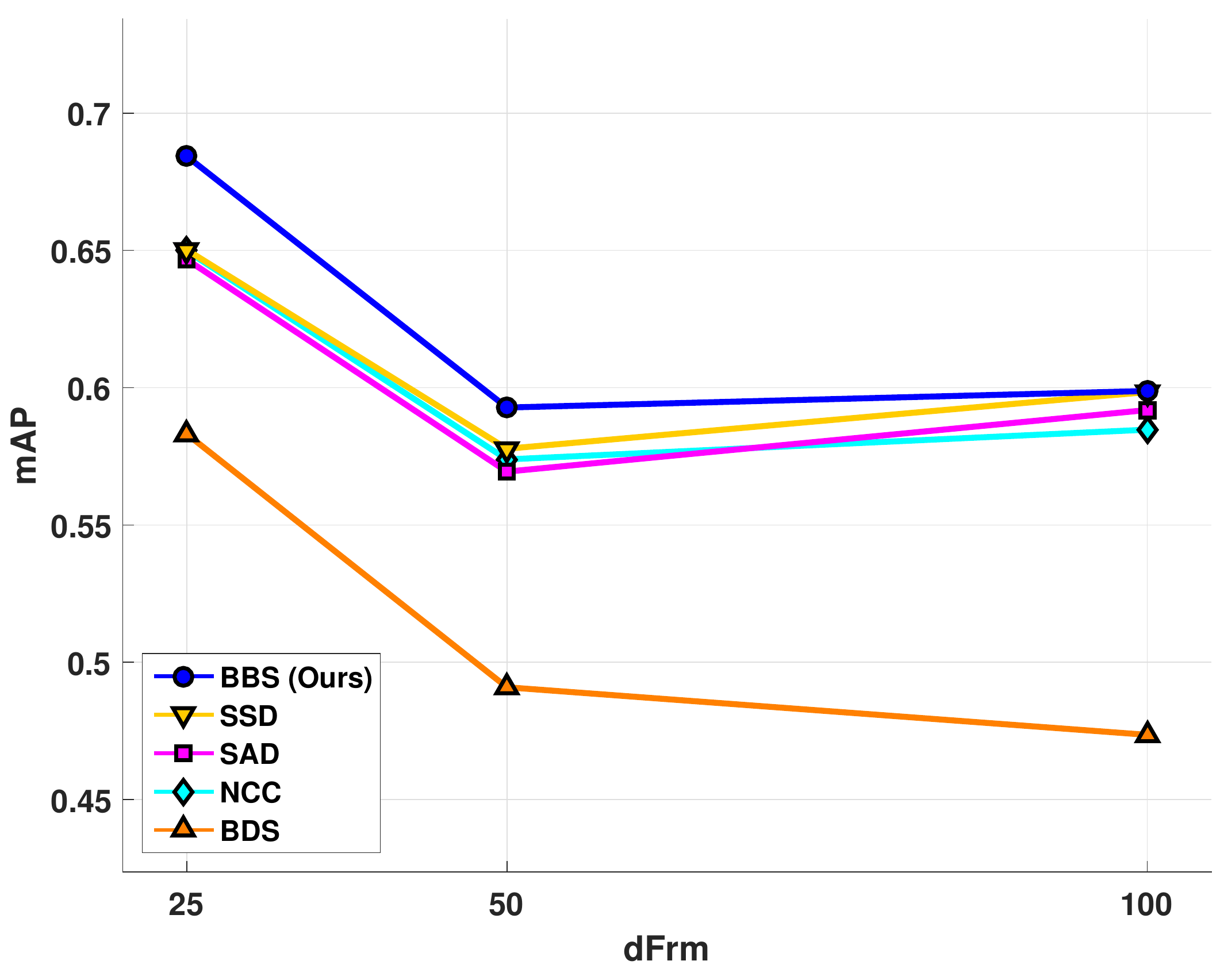} \\
		(a) Color features& (b) Deep features\\		
 	\end{tabular}
 	\caption{{\bf Effect of space time baseline:} Methods performance evaluated for data sets with different space-time baseline, $df=25,50$ and $100$. Left: Color features, Right: Deep features. BBS outperforms competing methods for both feature choices and for all $df$ values. Best viewed in color.}
 	\label{fig: mAP vs df} \vspace{-0.3cm}
 \end{figure*}

\subsection{Quantitative Evaluation} \label{sec:res_quan}
We now turn to the quantitative evaluation. The data for our experiments was generated from a dataset of 100 annotated video sequences\footnote{https://sites.google.com/site/benchmarkpami/}~\cite{Wu13}, both color and gray-scale. These videos capture a wide range of challenging scenes in which the objects of interest are diverse and typically undergo nonrigid deformations, photometric changes, motion blur, in/out-of-plane rotation, and occlusions.

Three template matching datasets were randomly sampled from the annotated videos. Each dataset is comprised of template-image pairs, where each such pair consists of frames $f$ and $f+df$, where $f$ was randomly chosen. For each dataset a different value of $df$ was used ($25,50$ or $100$). The ground-truth annotated bounding box in frame $f$ is used as the template, while frame $f+df$ is used as the query image.  This random choice of frames creates a challenging benchmark with a wide baseline in both time and space (see examples in  Figure ~\ref{fig: examples} and Figure ~\ref{fig: examples_df}). For $df = 25,50$ the data sets consist of $270$ pairs and for $df = 100$ there are $254$ pairs.

BBS using both color (with $HSV$ color space) and Deep features was compared with the 5 similarity measures mentioned above. The ground-truth annotations were used for quantitative evaluation. Specifically, we measure the accuracy of both the top match as well as the top k ranked matches, as follows.

\minesubsec{Accuracy:} was measured using the common bounding box overlap measure: $Acc.\!=\!\frac{\textrm{area}(B_{e}\cap B_{g})}{\textrm{area}(B_{e}\cup B_{g})}$ where $B_{e}$ and $B_{g}$ are the estimated and ground truth bounding boxes, respectively. The ROC curves show the fraction of examples with overlap larger than a threshold ($TH\in[0,1]$). Mean average precision (mAP) is taken as the area-under-curve (AUC). 
The success rates, of all methods, were evaluated considering only the global maximum (best mode) prediction as well as considering the best out of the top 3 modes (using non-maximum suppression, NMS). 

Results for both color feature and Deep features for the dataset with $df = 25$ are shown in Figure \ref{fig: acc}. Overall it can be seen that BBS outperforms competing methods using both color and Deep features. Using color features and considering only the top mode Figure \ref{fig: acc}(a), BBS outperforms competing methods with a margin ranging from $4.6\%$ compared to BDS to over $30\%$ compared to SSD. When considering the top 3 modes, Figure \ref{fig: acc}(c), the performance of all methods improves. However, we can clearly see the dominance of BBS, increasing its margin over competing methods. BBS reaches mAP of 0.648 (compared to 0.589 with only the top mode). For example the margin between BBS and BDS, which is the runner up, increases to $5.9\%$. The increase in performance when considering the top 3 modes suggests that there are cases where BBS is able to produce a mode at the correct target position however this mode might not be the global maximum of the entire map. 

Some successful template matching examples, along with the likelihood maps produced by BBS, using the color features, are shown in Figure ~\ref{fig: examples}. Notice how BBS can overcome non-rigid deformations of the target. 

\begin{figure*}[t!]
 	\centering
 	\begin{tabular}{cccc} 		
		\includegraphics[width = 0.23\textwidth, height = 0.15\textwidth]{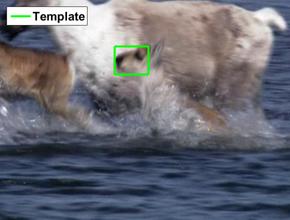} &
		\includegraphics[width = 0.23\textwidth, height = 0.15\textwidth]{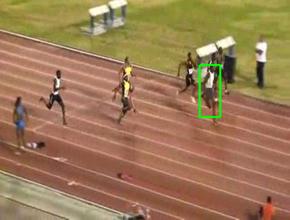} & 
		\includegraphics[width = 0.23\textwidth, height = 0.15\textwidth]{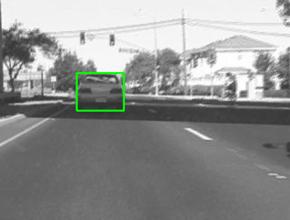} & 
		\includegraphics[width = 0.23\textwidth, height = 0.15\textwidth]{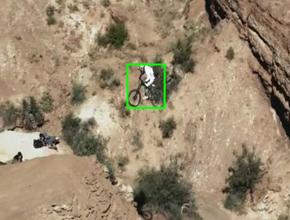} \\ 
		\includegraphics[width = 0.23\textwidth, height = 0.15\textwidth]{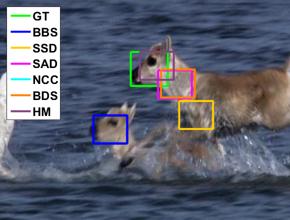} &
		\includegraphics[width = 0.23\textwidth, height = 0.15\textwidth]{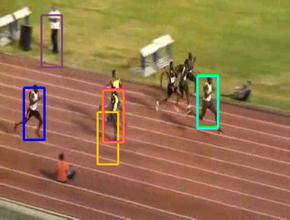} & 
		\includegraphics[width = 0.23\textwidth, height = 0.15\textwidth]{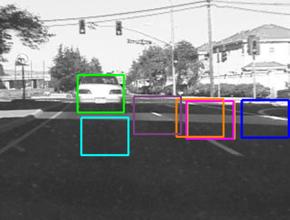} & 
		\includegraphics[width = 0.23\textwidth, height = 0.15\textwidth]{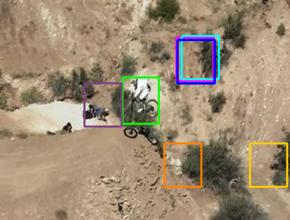} \\
		(a) & (b) & (c) & (d)\\				
 	\end{tabular}
 	\caption{{\bf Example of failure cases using color features}. Top, input images with annotated template marked in green.  Bottom,  target images and detected bounding boxes (see legend); ground-truth (GT) marked in green (our results in blue). As can be seen, some common failure causes are illumination changes, similar distracting targets or locking onto the background.}
 	\label{fig: fail} \vspace{-3mm}
 \end{figure*}

Typical failure cases are presented in Figure ~\ref{fig: fail}. Most of the failure cases using the color features can be attributed to either illumination variations (c), distracting objects with a similar appearance to the target (a)-(b), or cases were BBS matches the background or occluding object rather than the target (d). This usually happens when the target is heavily occluded or when the background region in the target window is very large. 

Results using our Deep feature and considering only the top mode are shown in figures Figure \ref{fig: acc}(b). We note that HM was not evaluated in this case due to the high dimensionality of the feature space. We observe that BBS outperforms the second best methods by only a small margin of $2.4\%$. Considering the top 3 modes allows BBS to reach mAP of 0.684 increasing its margin relative to competing methods. For example the margin relative to the second best method (SSD) is now $5.2\%$. 

Some template matching examples, along with their associated likelihood maps, using the Deep features, are shown in Figure ~\ref{fig: examples_df}. The Deep features are not sensitive to illumination variations and can capture both low level information as well as higher level object semantics. As can be seen the combination of using Deep features and BBS can deliver superior results due to its ability to explain non-rigid deformations. Note how when using the Deep feature, we can correctly match the bike rider in Figure ~\ref{fig: examples_df}(c) for which color features failed (Figure ~\ref{fig: fail}(d)). BBS with Deep features produce very well localized and compact modes compared to when color features are used.

\begin{figure*}[t!]
 	\centering
 	\begin{tabular}{cccc} 		
		\includegraphics[width = 0.23\textwidth, height = 0.15\textwidth]{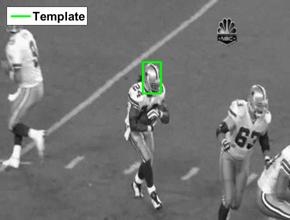} &
		\includegraphics[width = 0.23\textwidth, height = 0.15\textwidth]{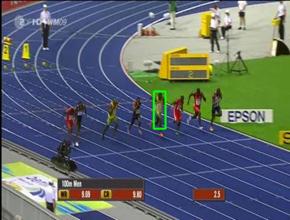} & 
		\includegraphics[width = 0.23\textwidth, height = 0.15\textwidth]{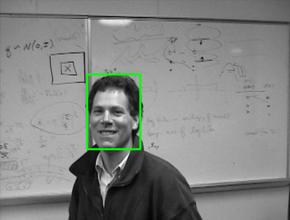} & 
		\includegraphics[width = 0.23\textwidth, height = 0.15\textwidth]{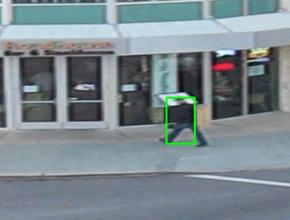} \\ 
		\includegraphics[width = 0.23\textwidth, height = 0.15\textwidth]{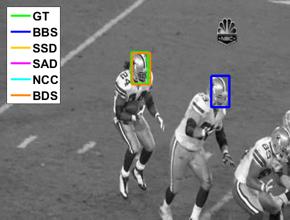} &
		\includegraphics[width = 0.23\textwidth, height = 0.15\textwidth]{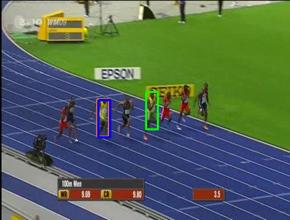} & 
		\includegraphics[width = 0.23\textwidth, height = 0.15\textwidth]{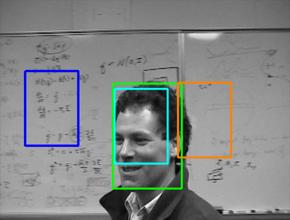} & 
		\includegraphics[width = 0.23\textwidth, height = 0.15\textwidth]{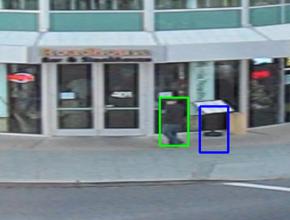} \\
		(a) & (b) & (c) & (d)\\						
 	\end{tabular}
 	\caption{{\bf Example of failure cases using Deep features}. Top, input images with annotated template marked in green.  Bottom,  target images and detected bounding boxes (see legend); ground-truth (GT) marked in green (our results in blue). Some common failure causes are similar distracting targets or locking onto the background.}
 	\label{fig: fail_df} \vspace{-3mm}
 \end{figure*}

Some typical failure cases when using the Deep features are presented in Figure ~\ref{fig: fail_df}. As for the color features, many failure cases are due to distracting objects with a similar appearance (a)-(b) or cases were BBS matches the background or occluding object (d).

It is interesting to see that BDS which was the runner up when color features were used come in last when using Deep features switching places with SSD which was worst previously and is now second in line. This also demonstrates the robustness of BBS which is able to successfully use different features. Additionally, we see that overall BBS with Deep features outperforms BBS with color features (a margin of $5.5\%$ with top 3 modes). However, this performance gain requires a significant increased in computational load both since the features have to be extracted and also since the proposed efficient computation scheme cannot be used in this case.
It is interesting to see that BBS with color features is able perform as well as SSD with Deep features.

Finally, we note that, when using the color features BBS outperforms HM which uses the $\chi^2$ distance. Although BBS converges to $\chi^2$ for large sets there are clear benefits for using BBS over $\chi^2$. Computing BBS does not require modeling the distributions (i.e. building normalized histograms) and can be performed on the raw data itself. This alleviates the need to choose the histogram bin size which is known to be a delicate issue. Moreover, BBS can be performed on high dimensional data, such as our Deep features, for which modeling the underlying distribution is not practical.

\minesubsec{The space time baseline:} effect on performance was examined using data-sets with different $df$ values ($25,50,100$). Figure \ref{fig: mAP vs df} shows mAP of competing methods for different values of $df$. Results using color features are shown on the left and using Deep features on the right. All results were analyzed taking the best out of the top 3 modes. It can be seen that BBS outperforms competing methods for the different $df$ values with the only exception being Deep feature with $df=100$ in which case BBS and SSD produce similar results reaching mAP of 0.6.  

	\section{Conclusions}
We have presented a novel similarity measure between sets of objects called the Best-Buddies Similarity (BBS). BBS leverages statistical properties of mutual nearest neighbors and was shown to be useful for template matching in the wild. Key features of BBS were identified and analyzed demonstrating its ability to overcome several challenges that are common in real life template matching scenarios. It was also shown, that for sufficiently large point sets, BBS converges to the Chi-Square distance. This result provides interesting insights into the statistical properties of mutual nearest neighbors, and the advantages of using BBS over $\chi^2$ where discussed. 

Extensive qualitative and quantitative experiments on challenging data were performed and a caching scheme allowing for an efficient computation of BBS was proposed. BBS was shown to outperform commonly used template matching methods such as normalized cross correlation, histogram matching and bi-directional similarity. 
Different types of features can be used with BBS, as was demonstrated in our experiments, where superior performance was obtained using both color features as well as Deep features. 

Our method may fail when the template is very small compared to the target image, when similar targets are present in the scene or when the outliers (occluding object or background clutter) cover most of the template. In some of these cases it was shown that BBS can predict the correct position (produce a mode) but non necessarily give it the highest score. 
 
Finally, we note that since BBS is generally defined between sets of objects it might have additional applications in computer-vision or other fields that could benefit from its properties. A natural future direction of research is to explore the use of BBS as an image similarity measure, for object localization or even for document matching.  
%
 %
 %
	
	\appendices
	{\small
	\section{Proof of Lemma~\ref{lemma:1}}\label{A:lemma_proof}
Because of independent sampling, all points in $Q$ have equal probability being the best buddy of $p$. From this we have:
\BQ
\begin{array}{ll}
Pr[bb(p_i=p;P,Q)] = &\\
&=\sum_{i=1}^N Pr(bb(p,q_i;P,Q)=1)\\ 
& = N \cdot Pr(bb(p,q;P,Q)),\label{eq:all_q_equal}
\end{array}
\EQ
where $q$ is a point from $Q$ and subscript is dropped for ease of description.

The probability that two points are best buddies is given by:
\BQ
\begin{array}{l}
Pr(bb(p_i=p,q;P,Q)) =\\
 (F_Q(p^-)\! + \! 1 \!- \!F_Q(p^+) )^{N-1} (F_P(q^-)\!+\!1\!-\!F_P(q^+))^{N-1}.\label{eq:expect_bbs}
\end{array}
\EQ
where $F_P(x)$ and $F_Q(x)$ denote CDFs of these two distributions, that is, $F_P(x)\!=\!\Pr\{p \leq x\}$.
And, $p^- \!=\!p-|p-q|$, $p^+\!=\!p + |p-q|$, and $q^{+}, q^-$ are similarly defined.
Combining~Eq.\ref{eq:all_q_equal} and Eq.~\ref{eq:expect_bbs}, the probability that $p_i$ has a best buddy equals to
\BQ
\begin{array}{l}
\limitN N \intq (F_Q(p^-)\! + \! 1 \!- \!F_Q(p^+) )^{N-1}\\
\cdot (F_P(q^-)\!+\!1\!-\!F_P(q^+))^{N-1}f_Q(q)dq.\label{eq:bbs_inside}
\end{array}
\EQ

We denote the signed distance between two points by $m=p-q$. Intuitively, because the density function are non-zero at any place,
when $N$ goes to infinity, the probability that two points $p\in P, q\in Q$ are BBP decreases rapidly as $m$ increases.
Therefore, we only need to consider the case when the distance between $p$ and $q$ is very small. Formally, for any positive $\mb$, changing the integration limits in Eq.~\ref{eq:bbs_inside} from $\intp$
to $\int_{q=p-\mb}^{p+\mb}$ does not change the result (see Claim 2 in the supplementary material).

Then let us break down $F_P(\cdot)$ and $F_Q(\cdot)$ in Eq.~\ref{eq:bbs_inside}. Given that the density functions $f_P(p)$ and $f_Q(q)$ are Lipschitz continuous (Condition 2 in Theorem~\ref{theorem:1}), we can assume that they take a constant value in the interval $[p^-, p^+]$, and $[q^-, q^+]$. That is,
\begin{align}
f_P(p^-) \approx f_P(p^+) \approx f_P(p) \notag \\
f_Q(q^-) \approx f_Q(q^+) \approx f_Q(q) \label{eq:approx_f}
\end{align}
And thus, the expression $F_Q(p^+)-F_Q(p^-)$ can be approximated as follows:
\BQ
\begin{array}{l}
F_Q(p^+)-F_Q(p^-) = \\
=\int_{p^-}^{p^+} f_Q(q)dq \approx f_Q(q) \cdot (p^+-p^-) = 2|m|\cdot f_Q(p). \label{eq:approx_F}
\end{array}
\EQ
Similarly, $F_p(q^+)-F_P(q^-)\approx 2|m|\cdot f_P(q)$. Note that this approximation can also be obtained using Taylor expansion on $F_p(q^+)$ and $F_p(q^-)$. At last, since $p$ and $q$ are very close to each other, we assume:
\BQ
f_Q(q) \approx f_Q(p) \label{eq:approx_p_q}.
\EQ

Plugging all these approximations (Eq.~\ref{eq:approx_F} and Eq.~\ref{eq:approx_p_q}) to Eq.~\ref{eq:bbs_inside} and replacing $q$ by $m$, we get:
\begin{align}
\text{Eq.~\ref{eq:bbs_inside}}
    =&\notag\\ &=\limitN N \int_{m=-\mb}^{\mb} (1-2|m|f_Q(p))^{N-1} \notag\\&\cdot (1-2|m|f_P(p))^{N-1}f_Q(p)dq ~\label{eq:ebbs_approx1}\\
 	=& f_Q(p) \limitN N \int_{m=-\mb}^{\mb} \Big(1-2(f_P(p)+f_Q(p))|m| +\notag\\& 4f_P(p)f_Q(p)m^2\Big)^{N-1} dm ~\label{eq:ebbs_approx2}\\
    =& f_Q(p) \limitN N \int_{m=-\mb}^{\mb}\Big(1-2(f_P(p)+f_Q(p))m \Big)^{N-1} dm.
		~\label{eq:ebbs_approx3}
\end{align}
It is worth mentioning that the approximated equality in Eq.~\ref{eq:approx_F} and Eq.~\ref{eq:approx_p_q} becomes restrict equality when $N$ goes to infinity (for the proof see Claim 3 in the supplementary material).
Also, since the distance between two points $m$ is very small, the second order term $4f_P(p)f_Q(p)m^2$ in Eq.~\ref{eq:ebbs_approx2} is negligible and is dropped in Eq.~\ref{eq:ebbs_approx3}
(for full justification see Claim 4 in the supplementary material).

At last, 
$\limitN N \int_{m=-\mb}^{\mb} (1-a|m|)^{N-1} dm = \frac{2}{a}$ (see Claim 1 in supplementary material). Thus Eq.~\ref{eq:ebbs_approx3} equals to:
\BQ
\frac{f_Q(p)}{f_P(p)+f_Q(p)}\label{eq:ebbs_approx4}
\EQ
which completes the proof of Lemma~\ref{lemma:1}.


	}
\vspace{-3mm}
	\subsection*{Acknowledgments.}\noindent This work was supported in part by an Israel Science Foun- dation grant 1556/10, National Science Foundation Robust Intelligence 1212849 Reconstructive Recognition, and a grant from Shell Research.
	{\small
		\bibliographystyle{IEEEtran}
		\bibliography{BBS_arxiv}

\begin{thebibliography}{10}
\providecommand{\url}[1]{#1}
\csname url@samestyle\endcsname
\providecommand{\newblock}{\relax}
\providecommand{\bibinfo}[2]{#2}
\providecommand{\BIBentrySTDinterwordspacing}{\spaceskip=0pt\relax}
\providecommand{\BIBentryALTinterwordstretchfactor}{4}
\providecommand{\BIBentryALTinterwordspacing}{\spaceskip=\fontdimen2\font plus
\BIBentryALTinterwordstretchfactor\fontdimen3\font minus
  \fontdimen4\font\relax}
\providecommand{\BIBforeignlanguage}[2]{{%
\expandafter\ifx\csname l@#1\endcsname\relax
\typeout{** WARNING: IEEEtran.bst: No hyphenation pattern has been}%
\typeout{** loaded for the language `#1'. Using the pattern for}%
\typeout{** the default language instead.}%
\else
\language=\csname l@#1\endcsname
\fi
#2}}
\providecommand{\BIBdecl}{\relax}
\BIBdecl

\bibitem{Dekel2015BBS}
T.~Dekel, S.~Oron, S.~Avidan, M.~Rubinstein, and W.~Freeman, ``Best buddies
  similarity for robust template matching,'' in \emph{Computer Vision and
  Pattern Recognition (CVPR), 2015 IEEE Conference on}.\hskip 1em plus 0.5em
  minus 0.4em\relax IEEE, 2015.

\bibitem{ouyang2012performance}
W.~Ouyang, F.~Tombari, S.~Mattoccia, L.~Di~Stefano, and W.-K. Cham,
  ``Performance evaluation of full search equivalent pattern matching
  algorithms,'' \emph{PAMI}, 2012.

\bibitem{Hel-OrHD14}
\BIBentryALTinterwordspacing
Y.~Hel{-}Or, H.~Hel{-}Or, and E.~David, ``Matching by tone mapping: Photometric
  invariant template matching,'' \emph{{IEEE} Trans. Pattern Anal. Mach.
  Intell.}, vol.~36, no.~2, pp. 317--330, 2014. [Online]. Available:
  \url{http://doi.ieeecomputersociety.org/10.1109/TPAMI.2013.138}
\BIBentrySTDinterwordspacing

\bibitem{elboher2013asymmetric}
E.~Elboher and M.~Werman, ``Asymmetric correlation: a noise robust similarity
  measure for template matching,'' \emph{Image Processing, IEEE Transactions
  on}, 2013.

\bibitem{chen2003fast}
J.-H. Chen, C.-S. Chen, and Y.-S. Chen, ``Fast algorithm for robust template
  matching with m-estimators,'' \emph{Signal Processing, IEEE Transactions on},
  2003.

\bibitem{sibiryakov2011fast}
A.~Sibiryakov, ``Fast and high-performance template matching method,'' in
  \emph{CVPR}, 2011.

\bibitem{shin2007fast}
B.~G. Shin, S.-Y. Park, and J.~J. Lee, ``Fast and robust template matching
  algorithm in noisy image,'' in \emph{Control, Automation and Systems, 2007.
  ICCAS'07. International Conference on}, 2007.

\bibitem{pele2008robust}
O.~Pele and M.~Werman, ``Robust real-time pattern matching using bayesian
  sequential hypothesis testing,'' \emph{PAMI}, 2008.

\bibitem{tsai2002rotation}
D.-M. Tsai and C.-H. Chiang, ``Rotation-invariant pattern matching using
  wavelet decomposition,'' \emph{Pattern Recognition Letters}, 2002.

\bibitem{kim2007grayscale}
H.~Y. Kim and S.~A. De~Ara{\'u}jo, ``Grayscale template-matching invariant to
  rotation, scale, translation, brightness and contrast,'' in
  \emph{AIVT}.\hskip 1em plus 0.5em minus 0.4em\relax Springer, 2007.

\bibitem{cvpr2013Fast_Match}
S.~Korman, D.~Reichman, G.~Tsur, and S.~Avidan, ``Fast-match: Fast affine
  template matching,'' in \emph{CVPR}, 2013.

\bibitem{tian2012globally}
Y.~Tian and S.~G. Narasimhan, ``Globally optimal estimation of nonrigid image
  distortion,'' \emph{IJCV}, 2012.

\bibitem{comaniciu2000real}
D.~Comaniciu, V.~Ramesh, and P.~Meer, ``Real-time tracking of non-rigid objects
  using mean shift,'' in \emph{CVPR}, 2000.

\bibitem{perez2002color}
P.~P{\'e}rez, C.~Hue, J.~Vermaak, and M.~Gangnet, ``Color-based probabilistic
  tracking,'' in \emph{ECCV 2002}, 2002.

\bibitem{Bao12}
C.~Bao, Y.~Wu, H.~Ling, and H.~Ji, ``Real time robust l1 tracker using
  accelerated proximal gradient approach,'' \emph{CVPR}, 2012.

\bibitem{Jia12}
X.~Jia, H.~Lu, and M.~Yang, ``Visual tracking via adaptive structural local
  sparse appearance model,'' \emph{CVPR}, 2012.

\bibitem{Olson02}
\BIBentryALTinterwordspacing
C.~F. Olson, ``Maximum-likelihood image matching,'' \emph{{IEEE} Trans. Pattern
  Anal. Mach. Intell.}, vol.~24, no.~6, pp. 853--857, 2002. [Online].
  Available:
  \url{http://doi.ieeecomputersociety.org/10.1109/TPAMI.2002.1008392}
\BIBentrySTDinterwordspacing

\bibitem{Oron2014LOT}
S.~Oron, A.~Bar-Hillel, D.~Levi, and S.~Avidan, ``Locally orderless tracking,''
  \emph{IJCV}, 2014.

\bibitem{Rubner00}
Y.~Rubner, C.~Tomasi, and L.~Guibas, ``The earth mover's distance as a metric
  for image retrieval,'' \emph{IJCV}, 2000.

\bibitem{Simakov2008summarizing}
D.~Simakov, Y.~Caspi, E.~Shechtman, and M.~Irani, ``Summarizing visual data
  using bidirectional similarity,'' in \emph{CVPR}, 2008.

\bibitem{huttenlocher1993comparing}
D.~P. Huttenlocher, G.~A. Klanderman, and W.~J. Rucklidge, ``Comparing images
  using the hausdorff distance,'' \emph{Pattern Analysis and Machine
  Intelligence, IEEE Transactions on}, vol.~15, no.~9, pp. 850--863, 1993.

\bibitem{Dubuisson94}
M.-P. Dubuisson and A.~Jain, ``A modified hausdorff distance for object
  matching,'' in \emph{Pattern Recognition, 1994. Vol. 1 - Conference A:
  Computer Vision amp; Image Processing., Proceedings of the 12th IAPR
  International Conference on}, vol.~1, Oct 1994, pp. 566--568 vol.1.

\bibitem{snedegor1967statistical}
G.~Snedegor, W.~G. Cochran \emph{et~al.}, ``Statistical methods.''
  \emph{Statistical methods.}, no. 6th ed, 1967.

\bibitem{Varma09}
M.~Varma and A.~Zisserman, ``A statistical approach to material classification
  using image patch exemplars,'' \emph{IEEE Transactions on Pattern Analysis
  and Machine Intelligence}, vol.~31, no.~11, pp. 2032--2047, 2009.

\bibitem{belongie2002shape}
S.~Belongie, J.~Malik, and J.~Puzicha, ``Shape matching and object recognition
  using shape contexts,'' \emph{Pattern Analysis and Machine Intelligence, IEEE
  Transactions on}, vol.~24, no.~4, pp. 509--522, 2002.

\bibitem{forssen2007shape}
P.-E. Forss{\'e}n and D.~G. Lowe, ``Shape descriptors for maximally stable
  extremal regions,'' in \emph{Computer Vision, 2007. ICCV 2007. IEEE 11th
  International Conference on}.\hskip 1em plus 0.5em minus 0.4em\relax IEEE,
  2007, pp. 1--8.

\bibitem{martin2004learning}
D.~R. Martin, C.~C. Fowlkes, and J.~Malik, ``Learning to detect natural image
  boundaries using local brightness, color, and texture cues,'' \emph{Pattern
  Analysis and Machine Intelligence, IEEE Transactions on}, vol.~26, no.~5, pp.
  530--549, 2004.

\bibitem{PomeranzSB11}
\BIBentryALTinterwordspacing
D.~Pomeranz, M.~Shemesh, and O.~Ben{-}Shahar, ``A fully automated greedy square
  jigsaw puzzle solver,'' in \emph{The 24th {IEEE} Conference on Computer
  Vision and Pattern Recognition, {CVPR} 2011, Colorado Springs, CO, USA, 20-25
  June 2011}, 2011, pp. 9--16. [Online]. Available:
  \url{http://dx.doi.org/10.1109/CVPR.2011.5995331}
\BIBentrySTDinterwordspacing

\bibitem{Li2015}
\BIBentryALTinterwordspacing
T.-t. Li, B.~Jiang, Z.-z. Tu, B.~Luo, and J.~Tang, \emph{Intelligent
  Computation in Big Data Era}.\hskip 1em plus 0.5em minus 0.4em\relax Berlin,
  Heidelberg: Springer Berlin Heidelberg, 2015, ch. Image Matching Using Mutual
  k-Nearest Neighbor Graph, pp. 276--283. [Online]. Available:
  \url{http://dx.doi.org/10.1007/978-3-662-46248-5_34}
\BIBentrySTDinterwordspacing

\bibitem{Liu2010}
H.~Liu, S.~Zhang, J.~Zhao, X.~Zhao, and Y.~Mo, ``A new classification algorithm
  using mutual nearest neighbors,'' in \emph{2010 Ninth International
  Conference on Grid and Cloud Computing}, Nov 2010, pp. 52--57.

\bibitem{Ozaki2011}
\BIBentryALTinterwordspacing
K.~Ozaki, M.~Shimbo, M.~Komachi, and Y.~Matsumoto, ``Using the mutual k-nearest
  neighbor graphs for semi-supervised classification of natural language
  data,'' in \emph{Proceedings of the Fifteenth Conference on Computational
  Natural Language Learning}, ser. CoNLL '11.\hskip 1em plus 0.5em minus
  0.4em\relax Stroudsburg, PA, USA: Association for Computational Linguistics,
  2011, pp. 154--162. [Online]. Available:
  \url{http://dl.acm.org/citation.cfm?id=2018936.2018954}
\BIBentrySTDinterwordspacing

\bibitem{Hu2012}
Z.~Hu and R.~Bhatnagar, ``Clustering algorithm based on mutual k-nearest
  neighbor relationships,'' \emph{Statistical Analy Data Mining}, vol.~5,
  no.~2, pp. 110--113, 2012.

\bibitem{Ma-ICCV-2015}
C.~Ma, J.-B. Huang, X.~Yang, and M.-H. Yang, ``Hierarchical convolutional
  features for visual tracking,'' in \emph{Proceedings of the IEEE
  International Conference on Computer Vision)}, 2015.

\bibitem{Wang_2015_ICCV}
L.~Wang, W.~Ouyang, X.~Wang, and H.~Lu, ``Visual tracking with fully
  convolutional networks,'' in \emph{The IEEE International Conference on
  Computer Vision (ICCV)}, December 2015.

\bibitem{ahonen2006face}
T.~Ahonen, A.~Hadid, and M.~Pietikainen, ``Face description with local binary
  patterns: Application to face recognition,'' in \emph{PAMI}, vol.~28,
  no.~12.\hskip 1em plus 0.5em minus 0.4em\relax IEEE, 2006, pp. 2037--2041.

\bibitem{Simonyan14c}
K.~Simonyan and A.~Zisserman, ``Very deep convolutional networks for
  large-scale image recognition,'' \emph{CoRR}, vol. abs/1409.1556, 2014.

\bibitem{Wu13}
Y.~Wu, J.~Lim, and M.~Yang, ``Online object tracking: A benchmark,'' in
  \emph{CVPR}, 2013.

\end{thebibliography}
	}
	\vspace{-3mm}

\end{document}